\documentclass[10pt,journal,compsoc]{IEEEtran}

\ifCLASSOPTIONcompsoc
  \usepackage[nocompress]{cite}
\else
  \usepackage{cite}
\fi

\ifCLASSINFOpdf
\else
\fi

\usepackage{times}
\usepackage{epsfig}
\usepackage{graphicx}
\usepackage{amsmath}
\usepackage{amssymb}
\usepackage{xcolor}
\usepackage{dsfont}

\usepackage[utf8]{inputenc} 
\usepackage[T1]{fontenc}    
\usepackage{url}            
\usepackage{booktabs}       
\usepackage{amsfonts}       
\usepackage{nicefrac}       
\usepackage{microtype}      
\usepackage{graphicx}
\usepackage{mathabx}
\usepackage{longtable}
\usepackage{amsthm}   
\usepackage{stackrel}  

\usepackage{bm} 
\usepackage{subfig}

\usepackage{float}  
\usepackage{multirow}
\usepackage{xcolor,colortbl} 
\usepackage{color}  
\usepackage[linesnumbered,ruled]{algorithm2e}  
\usepackage[skip=2pt]{caption}  

\DeclareCaptionFormat{myformat}{\small\selectfont#1#2#3}
\usepackage[pagebackref=true,breaklinks=true,letterpaper=true,colorlinks,bookmarks=false]{hyperref}

\providecommand{\realnum}{\mathbb{R}}
\providecommand{\naturalnum}{\mathbb{N}}
\providecommand{\bmcal}[1]{\bm{\mathcal{#1}}}

 \providecommand{\matnot}[1]{_{[{#1}]}}  
 \providecommand{\invar}{z}  
\providecommand{\binvar}{\bm{\invar}}  
\providecommand{\outvar}{x}  
\providecommand{\boutvar}{\bm{\outvar}}  
\providecommand{\citep}{\cite} 
\providecommand{\citet}{\cite}

\providecommand{\etal}{\emph{et al}.}

\newcommand{\modelname}{ProdPoly}
\newcommand{\modelone}{CCP}
\newcommand{\modeltwo}{NCP}
\newcommand{\modelthree}{NCP-Skip}
\newcommand{\resnet}{ResNet}
\newcommand{\modelres}{Prodpoly-\resnet}

\newcommand{\rebuttal}[1]{{#1}}

\begin{document}

\title{Deep Polynomial Neural Networks}

\author{
Grigorios G. Chrysos, \quad Stylianos Moschoglou, \quad Giorgos Bouritsas, \\ \quad Jiankang Deng, \quad Yannis Panagakis, \quad Stefanos Zafeiriou 
\IEEEcompsocitemizethanks{\IEEEcompsocthanksitem SM, GB, JD, SZ are with the Department
of Computing, Imperial College London, SW7 2AZ, UK. GC is with the Department of Electrical Engineering, Ecole Polytechnique Federale de Lausanne (EPFL), Switzerland. YP is with the Department of Informatics and Telecommunications , University of Athens, GR.

Corresponding author's e-mail: grigorios.chrysos@epfl.ch}
\thanks{}
}

\IEEEtitleabstractindextext{
\begin{abstract}
Deep Convolutional Neural Networks (DCNNs) are currently the method of choice both for generative, as well as for discriminative learning in computer vision and machine learning. The success of DCNNs can be attributed to the careful selection of their building blocks (e.g., residual blocks, rectifiers, sophisticated normalization schemes, to mention but a few). In this paper, we propose $\Pi$-Nets, \rebuttal{a new class of function approximators based on polynomial expansions}. $\Pi$-Nets are polynomial neural networks, i.e., the output is a high-order polynomial of the input. The unknown parameters, which are naturally represented by high-order tensors, are estimated through a collective tensor factorization with factors sharing. We introduce three tensor decompositions that significantly reduce the number of parameters and show how they can be efficiently implemented by hierarchical neural networks. We empirically demonstrate that $\Pi$-Nets are very expressive and they even produce good results without the use of non-linear activation functions in a large battery of tasks and signals, i.e., images, graphs, and audio. When used in conjunction with activation functions, $\Pi$-Nets produce state-of-the-art results in three  challenging tasks, i.e. image generation, face verification and 3D mesh representation learning. \rebuttal{The source code is available at \url{https://github.com/grigorisg9gr/polynomial_nets}.}
\end{abstract}

}

\maketitle

\IEEEdisplaynontitleabstractindextext

\IEEEpeerreviewmaketitle

\section{Introduction}
\label{sec:prodpoly_introduction}

Deep Convolutional Neural Networks (DCNNs)~\cite{lecun1998gradient, krizhevsky2012imagenet} have demonstrated impressive results in a number of tasks the last few years~\cite{krizhevsky2012imagenet, huang2017densely, miyato2018spectral}. Arguably, the careful selection of architectural pipelines, e.g. skip connections~\cite{he2016deep}, normalization schemes~\cite{ioffe2015batch} etc., is significant, however the core structure relies on compositional functions of linear and nonlinear operators. Both theoretical~\cite{arora2018convergence, ji2018minimax} and empirical studies reveal the limitations of the existing structure.

Recent empirical~\cite{karras2018style} and theoretical~\cite{jayakumar2020Multiplicative} results support that multiplicative interactions expand the classes of functions that can be approximated. Motivated by these findings, we study a new class of function approximators, which we coin $\Pi-$nets, where the output is a polynomial function of the input. Specifically, we model a vector-valued function $\bm{G}(\bm{z}): \realnum^{d} \to \realnum^{o}$ by a high-order multivariate polynomial of the input $\bm{z}$, whose unknown parameters are naturally represented by high-order tensors. The number of parameters required to accommodate all higher-order correlations of the input explodes with the desired order of the polynomial. To that end, we cast polynomial parameters estimation as a coupled tensor factorization~\citep{sidiropoulos2017tensor} that jointly factorizes all the polynomial parameters tensors. We introduce three joint decompositions with shared factors and exhibit the resulting hierarchical structures (i.e., architectures of neural networks).

\begin{figure}[!t]
    \centering
    \includegraphics[width=1\linewidth]{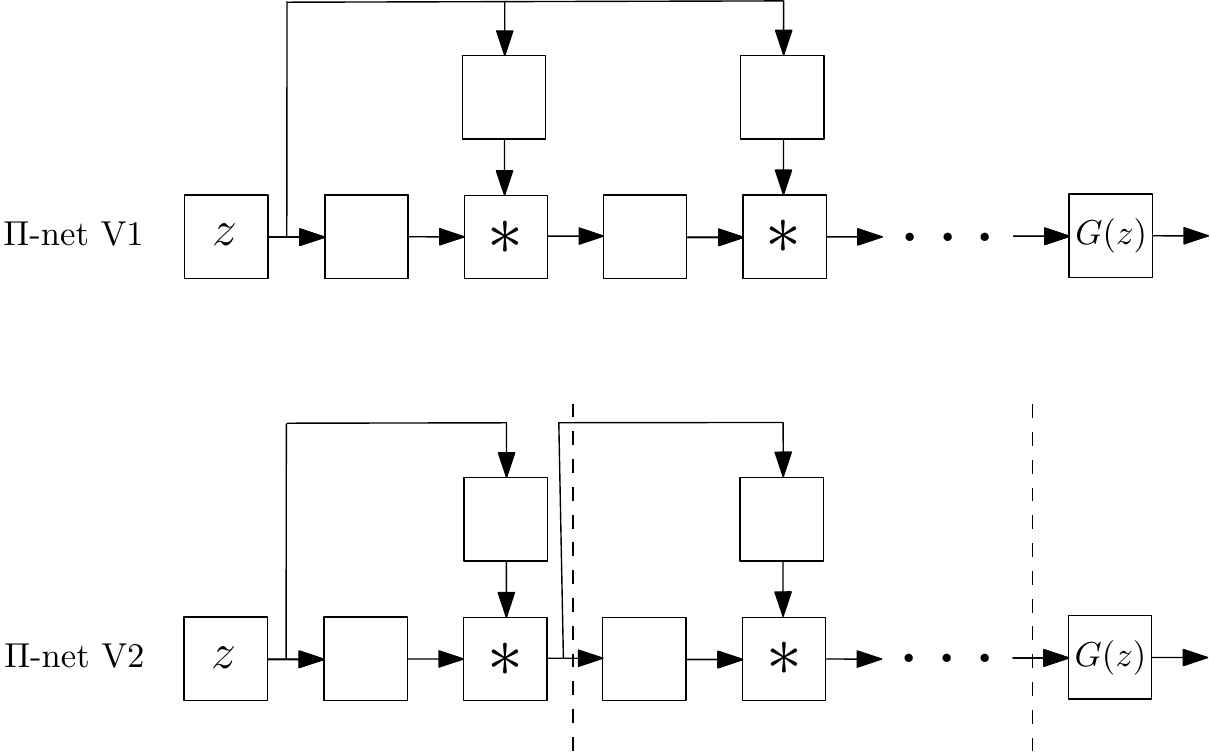}
\caption{In this paper we introduce a class of networks called $\Pi-$nets, where the output is a polynomial of the input. The input in this case, $\binvar$, can be either the latent space of Generative Adversarial Network for a generative task or an image in the case of a discriminative task. Our polynomial networks can be easily implemented.}
\label{fig:prodpoly_model_intro_schematic}
\end{figure}

In our preliminary works~\cite{chrysos2019polygan, chrysos2019newton, chrysos2020poly}, we introduced the concept of higher-order expansions for both generative and discriminative networks. In this work, our improvements are threefold. The concepts and the motivation behind each model are elaborated; the new intuitions will enable practitioners to devise new models tailored to their specific tasks. In addition, we extend the experimental results, e.g. include experiment in the challenging task of face verification and identification. Lastly, we conduct a thorough discussion on several challenging topics that require further work on this new class of neural networks. 

In particular, the paper bears the following contributions:
\begin{itemize}
    \item A new family of neural networks (called $\Pi-$nets) where the output is a high-order polynomial of the input is introduced. To avoid the combinatorial explosion in the number of parameters of polynomial activation functions~\citep{kileel2019expressive}, our $\Pi-$nets cast polynomial parameters estimation as a coupled tensor factorization with shared factors (please see Fig.~\ref{fig:prodpoly_model_intro_schematic} for an indicative schematic representation).
    \item The proposed architectures are applied in a) generative models such as GANs, and b) discriminative networks. Additionally, the polynomial architectures are used to learn high-dimensional distributions without non-linear activation functions. 
    \item We convert state-of-the-art baselines using the proposed $\Pi-$nets and show how they can largely improve the performance of the baseline. We demonstrate it conclusively in a battery of tasks (i.e., generation, classification and face verification/identification). Our architectures are applicable to many different signals (e.g. images, meshes, and audio) and outperform the prior art.
\end{itemize}

The rest of the paper is organized as follows: Sec.~\ref{sec:prodpoly_related} summarizes the related work. In Sec.~\ref{sec:prodpoly_method} we introduce the polynomial networks and showcase the resulting architectures for three decompositions. \rebuttal{The core experimental evaluation is conducted in Sec.~\ref{sec:prodpoly_experiments}, while a number of experiments are deferred to the supplementary.} The existing limitations and future directions of the polynomial networks are discussed in Sec.~\ref{sec:prodpoly_discussion}, while Sec.~\ref{sec:prodpoly_conclusion} concludes the paper.

 \section{Related work and notation}
\label{sec:prodpoly_related}

\textbf{Expressivity of (deep) neural networks}: The last few years, (deep) neural networks have been applied to a wide range of applications with impressive results. The performance boost can be attributed to a host of factors including: a) the availability of massive datasets~\cite{deng2009imagenet, liu2015deep}, b) the machine learning libraries~\cite{chainer_learningsys2015, paszke2017automatic} running on massively parallel hardware, c) training improvements. The training improvements include a) optimizer improvement~\cite{kingma2014adam, reddi2019convergence}, b) augmented capacity of the network~\cite{simonyan2014very}, c) regularization tricks~\cite{glorot2010understanding, saxe2013exact, ioffe2015batch, ulyanov2016instance}. However, the paradigm for each layer remains largely unchanged for several decades: each layer is composed of a linear transformation and an element-wise activation function. Despite the variety of linear transformations~\cite{fukushima1980neocognitron, lecun1998gradient, krizhevsky2012imagenet} and activation functions~\cite{ramachandran2017searching, nair2010rectified} being used, the effort to extend this paradigm has not drawn much attention to date.

Recently, hierarchical models have exhibited stellar performance in learning expressive generative models~\cite{brock2019large, karras2018style, zhao2017learning}. For instance, the recent BigGAN \citep{brock2019large} performs a hierarchical composition through skip connections from the noise $\bm{z}$ to multiple resolutions of the generator. A similar idea emerged in StyleGAN~\citep{karras2018style}, which is an improvement over the Progressive Growing of GANs (ProGAN)~\citep{karras2017progressive}. As ProGAN, StyleGAN is a highly-engineered network that achieves compelling results on synthesized 2D images. In order to provide an explanation on the improvements of StyleGAN over ProGAN, the authors adopt arguments from the style transfer literature \citep{huang2017arbitrary}. We believe that these improvements can be better explained under the light of our proposed polynomial function approximation. Despite the hierarchical composition proposed in these works, we present an intuitive and mathematically elaborate method to achieve a more precise approximation with a polynomial expansion. We also demonstrate that such a polynomial expansion can be used in both image generation (as in \cite{karras2018style, brock2019large}), image classification, and graph representation learning.

\begin{table*}[h]
\begin{minipage}{.7\linewidth}
    \caption{Nomenclature}
    \label{tbl:prodpoly_primary_symbols}
    \centering
    \begin{tabular}{|c | c | c|}
    \toprule
    Symbol 	& Dimension(s) 		&	Definition \\
    \midrule
    $n, N$ 		            & $\naturalnum$		            &	Polynomial term order, total approximation order. \\
    $k$ 		            & $\naturalnum$		            & Rank of the decompositions. \\
    $\binvar$            & $\realnum^d$                      & Input to the polynomial approximator. \\
    $\bm{C}, \bm{\beta}$ 		    & $\realnum^{o\times k}, \realnum^{o}$		        &	Parameters in all decompositions. \\
    $\bm{A}\matnot{n}, \bm{S}\matnot{n}, \bm{B}\matnot{n}$          &       $\realnum^{d\times k}, \realnum^{k\times k}, \realnum^{\omega\times k}$     & Matrix parameters in the hierarchical decomposition.\\
    $\odot, *$          &   -       & Khatri-Rao product, Hadamard product. \\
     \hline
    \end{tabular}
    
\end{minipage}
\begin{minipage}{.2\linewidth}
    \caption{Single polynomial models (Sec.~\ref{ssec:prodpoly_single_poly})}
    \label{tbl:prodpoly_single_polynomial_models}
    \centering
    \begin{tabular}{|c | c | c |}
    \toprule
    Name 	& Schematic 	& Recursive eq.\\
    \midrule
    \modelone 		            & Fig.~\ref{fig:prodpoly_model1_schematic}     & \eqref{eq:prodpoly_model1}\\
    \modeltwo 		            & Fig.~\ref{fig:prodpoly_model2_schematic}     & \eqref{eq:prodpoly_model2}\\
    \modelthree 		            & Fig.~\ref{fig:prodpoly_model3_schematic} & \eqref{eq:prodpoly_model3}    \\
     \hline
    \end{tabular}
\end{minipage}
\end{table*}

\textbf{Polynomial networks}: Polynomial relationships have been investigated in two specific categories of networks: a) self-organizing networks with hard-coded feature selection, b) pi-sigma networks.

The idea of learnable polynomial features can be traced back to Group Method of Data Handling (GMDH)~\cite{ivakhnenko1971polynomial}\footnote{This is often referred to as the first deep neural network~\cite{schmidhuber2015deep}.}. GMDH learns partial descriptors that capture quadratic correlations between two predefined input elements. In \cite{oh2003polynomial}, more input elements are allowed, while higher-order polynomials are used. The input to each partial descriptor is predefined (subset of the input elements), which does not allow the method to scale to high-dimensional data with complex correlations.

Shin \etal~\cite{shin1991pi} introduce the pi-sigma network, which is a neural network with a single hidden layer. Multiple affine transformations of the data are learned; a product unit multiplies all the features to obtain the output. Improvements in the pi-sigma network include regularization for training in~\cite{xiong2007training} or using multiple product units to obtain the output in~\cite{voutriaridis2003ridge}. 
The pi-sigma network is extended in sigma-pi-sigma neural network (SPSNN)~\cite{li2003sigma}. The idea of SPSNN relies on summing different pi-sigma networks to obtain each output. SPSNN also uses a predefined basis (overlapping rectangular pulses) on each pi-sigma sub-network to filter the input features. Even though such networks use polynomial features or products, they do not scale well in high-dimensional signals. In addition, their experimental evaluation is conducted only on signals with known ground-truth distributions (and with up to 3 dimensional input/output), unlike the modern generative models where only a finite number of samples from high-dimensional ground-truth distributions is available. 

\rebuttal{Convolutional arithmetic circuits (ConvACs) are also related to our work. Arithmetic circuits are networks with two types of nodes: sum nodes (weighted sum of their inputs), and product nodes (computing the product of their inputs). Those two types of nodes are sufficient to express a polynomial expansion. On \cite{cohen2016expressive}, the authors want to characterize the depth efficiency of (deep) convolutional neural networks. The CP decomposition is used to factorize the weights of a shallow convolutional network, while the hierarchical Tucker decomposition is used for the deep network. In \cite{cohen2016convolutional}, the authors generalize their previous results by inserting nonlinear activation functions (only linear activation functions were considered in \cite{cohen2016expressive}). The aforementioned works have a complementary role to our work, since their intent is to characterize (the depth efficiency of) convolutional neural networks, while our goal is to use polynomial expansion to approximate the target function.}

Another instance of such polynomial networks is through multiplicative interactions. Recently, there is a surge of methods~\citep{bahdanau2014neural, srivastava2015highway, reed2014learning} reporting superior performance through multiplicative interactions. The work of \citet{jayakumar2020Multiplicative} provides a theoretical understanding on why such connections might be beneficial. The aforementioned works model interactions of second or third order. Polynomial networks can be seen as high-order generalizations of such multiplicative interactions~\citet{jayakumar2020Multiplicative, bahdanau2014neural, srivastava2015highway}.

\subsection{\rebuttal{Notation}}
\rebuttal{Tensors\footnote{Further details on the tensor notation are deferred to the supplementary.} are symbolized by calligraphic letters, e.g.,  $\bmcal{X}$, while matrices (vectors) are denoted by uppercase (lowercase) boldface letters e.g., $\bm{X}$, ($\bm{x}$). A set of $M$ real matrices (vectors) of varying dimensions is denoted by $\{\bm{X}\matnot{m} \in \realnum^{I_m \times N} \}_{m=1}^M$ $( \{ \bm {x}\matnot{m} \in \realnum^{I_m} \}_{m=1}^M )$.  }

\rebuttal{\textbf{Products}: The \textit{Hadamard} product of $\bm{A} \in \realnum^{I \times N}$
and $\bm{B} \in \realnum^{I \times N}$ is defined as $\bm{A} * \bm{B}$ and is equal to ${A}_{(i, j)} {B}_{(i, j)}$ for the $(i, j)$ element. The \textit{Khatri-Rao} product of matrices $\bm{A} \in \realnum^{I \times N}$
and $\bm{B} \in \realnum^{J \times N}$ is
denoted by $\bm{A} \odot \bm{B}$ and yields a matrix of
dimensions $(IJ)\times N$.  For a set of matrices  $\{\bm{A}\matnot{m} \in \realnum^{I_m \times N} \}_{m=1}^M$ the Khatri-Rao product  is denoted by:}
\begin{equation}
    \bm{A}\matnot{1} \odot \bm{A}\matnot{2} \odot  \cdots \odot  \bm{A}\matnot{M} \doteq  \bigodot_{m=1}^M \bm{A}\matnot{m}
\end{equation} 

\rebuttal{\textbf{Tensors}: Each element of an $M^{th}$ order tensor $\bmcal{X}$ is addressed by $M$ indices, i.e., $(\bmcal{X})_{i_{1}, i_{2}, \ldots, i_{M}} \doteq x_{i_{1}, i_{2}, \ldots, i_{M}}$. An $M^{th}$-order  real-valued tensor $\bmcal{X}$ is  defined over the
tensor space $\realnum^{I_{1} \times I_{2} \times \cdots \times
I_{M}}$, where $I_{m} \in \mathbb{Z}$ for $m=1,2,\ldots,M$. 
The \textit{mode-$m$ vector product} of $\bmcal{X}$ with a
vector $\bm{u} \in \realnum^{I_m}$, denoted by
$\bmcal{X} \times_{m} \bm{u} \in \realnum^{I_{1}\times
I_{2}\times\cdots\times I_{m-1}  \times I_{m+1} \times
\cdots \times I_{M}} $, results in a tensor of order $M-1$:}
\begin{equation}\label{E:Tensor_Mode_n}
(\bmcal{X} \times_{m} \bm{u})_{i_1, \ldots, i_{m-1}, i_{m+1},
\ldots, i_{M}} = \sum_{i_m=1}^{I_m} x_{i_1, i_2, \ldots, i_{M}} u_{i_m}.
\end{equation}
\rebuttal{Furthermore,  we denote
$\bmcal{X} \times_{1} \bm{u}^{(1)} \times_{2} \bm{u}^{(2)} \times_{3}  \cdots \times_{M} \bm{u}^{(M)}  \doteq 
\bmcal{X} \prod_{m=1}^m \times_{m} \bm{u}^{(m)}$.}

A core tool in our analysis is the CP decomposition that factorizes a tensor into a sum of component rank-one tensors~\cite{kolda2009tensor}. By considering the mode-$1$ unfolding of an $M^{th}$-order tensor $\bmcal{X}$, the CP decomposition can be written in matrix form as \citep{kolda2009tensor}: $ \bm{X}_{(1)}  \doteq \bm{U}\matnot{1} \bigg( \bigodot_{m = M}^{2} \bm{U}\matnot{m}\bigg)^T$ where $\{ \bm{U}\matnot{m}\}_{m=1}^{M}$ are the factor matrices.

 \section{Method}
\label{sec:prodpoly_method}

We want to learn a function approximator where each element of the output $\outvar_j$, with $j\in [1, o]$, is expressed as a polynomial\footnote{The theorem of \citep{stone1948generalized} guarantees that any smooth function can be approximated by a polynomial. The approximation of multivariate functions is covered by an extension of the Weierstrass theorem, e.g., in \cite{nikol2013analysis} (pg 19).} of all the input elements $\invar_i$, with $i\in [1, d]$. That is, we want to learn a function $G: \realnum^{d} \to \realnum^{o}$ of order $N \in \naturalnum$, such that:

\begin{equation}
\begin{split}
    \outvar_j = G(\binvar)_j = \beta_j + {\bm{w}_j^{[1]}}^T\binvar + \binvar^T \bm{W}_j^{[2]}\binvar + \\
    \bmcal{W}_j^{[3]}\times_1\binvar\times_2\binvar\times_3\binvar + \cdots + \bmcal{W}_j^{[N]}\prod_{n=1}^N \times_{n} \binvar
\end{split}
\label{eq:prodpoly_starting_poly_eq_element}
\end{equation}

where $\beta_j \in \realnum$, and  $\big\{\bmcal{W}_j^{[n]} \in \realnum^{\prod_{m=1}^n\times_m d}\big\}_{n=1}^N$ are parameters for approximating the output $\outvar_j$. The correlations (of the input elements $\invar_i$) up to $N^{th}$ order emerge in \eqref{eq:prodpoly_starting_poly_eq_element}. 
A more compact expression of \eqref{eq:prodpoly_starting_poly_eq_element} is obtained by vectorizing the outputs:

\begin{equation}
    \boutvar = G(\binvar) = \sum_{n=1}^N \bigg(\bmcal{W}^{[n]} \prod_{j=2}^{n+1} \times_{j} \binvar\bigg) + \bm{\beta}
    \label{eq:prodpoly_poly_general_eq}
\end{equation}

where $\bm{\beta} \in \realnum^o$ and $\big\{\bmcal{W}^{[n]} \in  \realnum^{o\times \prod_{m=1}^{n}\times_m d}\big\}_{n=1}^N$ are the learnable parameters. This form of \eqref{eq:prodpoly_poly_general_eq} allows us to approximate any smooth function (for large $N$), however the parameters grow with $\mathcal{O}(d^N)$. 

A variety of methods, such as pruning~\cite{frankle2018lottery, han2015learning}, special linear operators~\cite{ding2017c} with reduced parameters, parameter sharing/prediction~\cite{yunpeng2017sharing, denil2013predicting}, can be employed to reduce the parameters. The aforementioned approaches are post-processing techniques, i.e., given a (pre-trained) network, they reduce the parameters of the specific network. Instead, we design two principled ways which allow an efficient implementation. The first method relies on performing an off-the-shelf tensor decomposition on \eqref{eq:prodpoly_poly_general_eq}, while the second considers the final polynomial as the product of lower-degree polynomials.

\subsection{Single polynomial}
\label{ssec:prodpoly_single_poly}
A tensor decomposition on the parameters is a natural way to reduce the parameters and to implement \eqref{eq:prodpoly_poly_general_eq} with a neural network. Below, we demonstrate how three such decompositions result in novel architectures for a neural network training. The main symbols are summarized in Table~\ref{tbl:prodpoly_primary_symbols}, while the equivalence between the recursive relationship and the polynomial is analyzed in the supplementary.

\vspace{3mm}
\noindent \textbf{Model 1: \modelone{} (Coupled CP decomposition)}

Instead of factorizing each parameter tensor $\bmcal{W}^{[n]}$ individually we propose to jointly factorize all the parameter tensors using a coupled CP decomposition~\cite{kolda2009tensor} with a specific pattern of factor sharing. To illustrate the factorization, we assume a third order approximation ($N=3$), and then provide the recursive relationship that can scale to arbitrary expansion. 

Let us assume that the parameter tensors admit the following coupled CP decomposition with the factors corresponding to lower-order levels of approximation being shared across all parameters tensors. That is: 
\begin{itemize}
    \item Let $\bm{W}^{[1]} = \bm{C}\bm{U}\matnot{1}^T$, be the parameters for first level of approximation.
    \item Let $\bmcal{W}^{[2]}$ being a superposition of of two weights tensors, namely $\bmcal{W}^{[2]} = \bmcal{W}^{[2]}_{1:2} + \bmcal{W}^{[2]}_{1:3}$, with $\bmcal{W}^{[2]}_{i:j}$ denoting parameters associated with the second order interactions across the $i^{th}$ and $j^{th}$ order of approximation. By enforcing the CP decomposition of the above tensors to share the factor with tensors corresponding to lower-order of approximation we obtain in matrix form: $\bm{W}^{[2]}_{(1)} = \bm{C}(\bm{U}\matnot{3} \odot \bm{U}\matnot{1})^T + \bm{C}(\bm{U}\matnot{2} \odot \bm{U}\matnot{1})^T$.
    \item Similarly, we enforce the third-order parameters tensor to admit the following CP decomposition (in matrix form) $\bm{W}^{[3]}_{(1)} = \bm{C}(\bm{U}\matnot{3} \odot \bm{U}\matnot{2} \odot \bm{U}\matnot{1})^T $. Note that all but the $\bm{U}\matnot{3}$ factor matrices are shared in the factorization of tensors capturing polynomial parameters for the first and second order of approximation.
\end{itemize}

The parameters are $\bm{C} \in \realnum^{o\times k}, \bm{U}\matnot{m} \in  \realnum^{d\times k}$ for $m=1,2,3$. Then, (\ref{eq:prodpoly_poly_general_eq}) for $N=3$ is written as:

\begin{equation}
\begin{split}
    & G(\bm{z}) = \bm{\beta} + \bm{C}\bm{U}\matnot{1}^T\bm{z} + \bm{C}\Big(\bm{U}\matnot{3} \odot \bm{U}\matnot{1}\Big)^T(\bm{z} \odot \bm{z}) + \\
    & \bm{C}\Big(\bm{U}\matnot{2} \odot \bm{U}\matnot{1}\Big)^T(\bm{z} \odot \bm{z}) + \\
    & \bm{C}\Big(\bm{U}\matnot{3} \odot \bm{U}\matnot{2} \odot \bm{U}\matnot{1}\Big)^T(\bm{z} \odot \bm{z} \odot \bm{z})
\label{eq:polygan_recursive_gen_third_order}
\end{split}
\end{equation}

\begin{figure}[!h]
    \centering
    \includegraphics[width=1\linewidth]{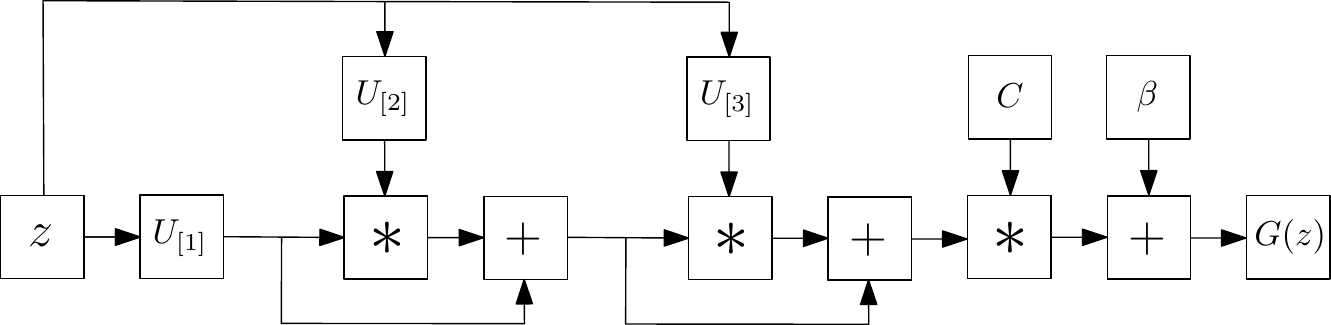}
\caption{Schematic illustration of the \modelone{} (for third order approximation). Symbol $*$ refers to the Hadamard product.}
\label{fig:prodpoly_model1_schematic}
\end{figure}

Using the Lemma 1 (provided in the supplementary), we can transform the \eqref{eq:polygan_recursive_gen_third_order} into a neural network as depicted in Fig.~\ref{fig:prodpoly_model1_schematic}. 

The \modelone{} factorization generalizes to $N^{th}$ order expansion. The recursive relationship for the $N^{th}$ order approximation is:

\begin{equation}
    \boutvar_{n} = \Big(\bm{U}\matnot{n}^T \binvar \Big)* \boutvar_{n-1} + \boutvar_{n-1}
    \label{eq:prodpoly_model1}
\end{equation}
for $n=2,\ldots,N$ with $\boutvar_{1} = \bm{U}\matnot{1}^T \binvar$ and $\boutvar = \bm{C}\boutvar_{N} + \bm{\beta}$. The parameters $\bm{C} \in \realnum^{o\times k}, \bm{U}\matnot{n} \in  \realnum^{d\times k}$ for $n=1,\ldots,N$ are learnable.

\noindent\textbf{Model 2: \modeltwo{} (Nested coupled CP decomposition)}

Instead of explicitly separating the interactions between layers, we can utilize a joint hierarchical decomposition on the polynomial parameters. Let us first introduce learnable hyper-parameters $\big\{\bm{b}\matnot{n} \in \realnum^\omega\big\}_{n=1}^N$, which act as scaling factors for each parameter tensor. Therefore, we modify (\ref{eq:prodpoly_poly_general_eq}) to: 

\begin{equation}
    G(\bm{z}) = \sum_{n=1}^N \bigg(\bmcal{W}^{[n]} \times_2 \bm{b}\matnot{N+1-n} \prod_{j=3}^{n+2} \times_{j} \bm{z}\bigg) + \bm{\beta},
    \label{eq:prodpoly_general_polynomial_with_b}
\end{equation}

with $\big\{\bmcal{W}^{[n]} \in  \realnum^{o\times \omega \times \prod_{m=1}^{n}\times_m d}\big\}_{n=1}^N$. Similarly to \modelone{}, we demonstrate the decomposition assuming a third order approximation ($N=3$), and then provide the general recursive relationship.

\begin{figure}[!h]
    \centering
    \includegraphics[width=1\linewidth]{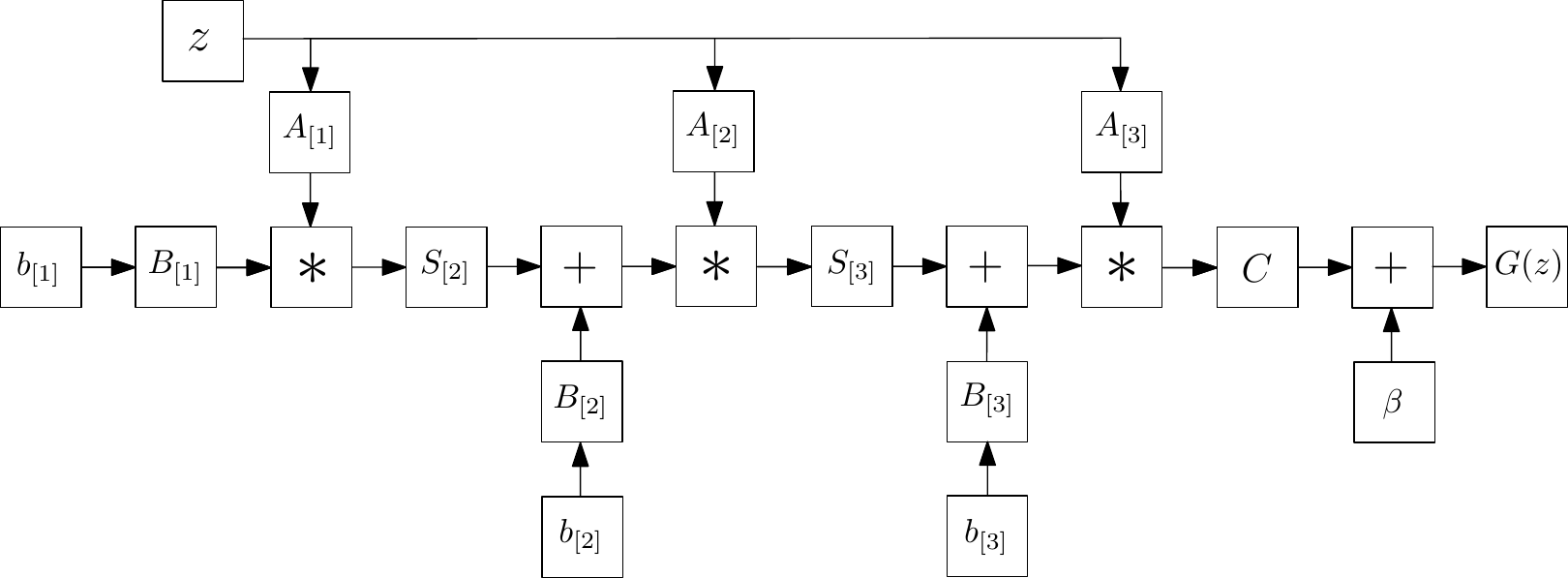}
\caption{Schematic illustration of the \modeltwo{} (for third order approximation). Symbol $*$ refers to the Hadamard product.}
\label{fig:prodpoly_model2_schematic}
\end{figure}

To estimate the parameters (in $N=3$ expansion) we jointly factorize all parameter tensors by employing nested CP decomposition with parameter sharing as follows (in matrix form):

\begin{itemize}
    \item First order parameters : $\bm{W}^{[1]}_{(1)} = \bm{C} (\bm{A}\matnot{3} \odot \bm{B}\matnot{3})^T$.
    \item Second order parameters: \\
    $\bm{W}^{[2]}_{(1)} = \bm{C} \bigg\{\bm{A}\matnot{3} \odot \bigg[\Big(\bm{A}\matnot{2} \odot \bm{B}\matnot{2}\Big) \bm{S}\matnot{3}\bigg]\bigg\}^T$.
    \item Third order parameters: \\
    $\bm{W}^{[3]}_{(1)} = \bm{C} \bigg\{\bm{A}\matnot{3} \odot \bigg[\bigg(\bm{A}\matnot{2} \odot \Big\{\Big(\bm{A}\matnot{1} \odot \bm{B}\matnot{1}\Big) \bm{S}\matnot{2}\Big\} \bigg)\bm{S}\matnot{3} \bigg]\bigg\}^T$
\end{itemize}

with $\bm{C} \in  \realnum^{o\times k}, \bm{A}\matnot{n} \in  \realnum^{d\times k}, \bm{S}\matnot{n} \in  \realnum^{k\times k}, \bm{B}\matnot{n} \in  \realnum^{\omega\times k}$ for $n=1,\ldots,N$.  
Altogether, (\ref{eq:prodpoly_general_polynomial_with_b}) for $N=3$ is written as:

\begin{equation}
\begin{split}
    & G(\bm{z}) = \bm{\beta} + \bm{C} (\bm{A}\matnot{3} \odot \bm{B}\matnot{3})^T (\bm{z} \odot \bm{b}\matnot{3}) + \\
    & \bm{C} \bigg\{\bm{A}\matnot{3} \odot \bigg[\Big(\bm{A}\matnot{2} \odot \bm{B}\matnot{2}\Big) \bm{S}\matnot{3}\bigg]\bigg\}^T \Big(\bm{z} \odot \bm{z} \odot \bm{b}\matnot{2} \Big) + \\
    & \bm{C} \bigg\{\bm{A}\matnot{3} \odot \bigg[\bigg(\bm{A}\matnot{2} \odot \Big\{\Big(\bm{A}\matnot{1} \odot \bm{B}\matnot{1}\Big) \bm{S}\matnot{2}\Big\} \bigg)\bm{S}\matnot{3} \bigg]\bigg\}^T \bm{\mu} 
    \label{eq:polygan_third_order_decomp_init}
\end{split}
\end{equation}

with $\bm{\mu} = \Big(\bm{z} \odot \bm{z} \odot \bm{z} \odot \bm{b}\matnot{1}\Big)$. Using Lemma1 and further algebraic operations (see Sec. 3.2 in the supplementary), \eqref{eq:polygan_third_order_decomp_init} can be implemented by a neural network as depicted in Fig.~\ref{fig:prodpoly_model2_schematic}.

The recursive relationship for $N^{th}$ order approximation is defined as:
\begin{equation}
    \boutvar_{n} = \Big(\bm{A}\matnot{n}^T\binvar\Big) * \Big(\bm{S}\matnot{n}^T \boutvar_{n-1} + \bm{B}\matnot{n}^T\bm{b}\matnot{n}\Big)
    \label{eq:prodpoly_model2}
\end{equation}
for $n=2,\ldots,N$ with $\boutvar_{1} = \Big(\bm{A}\matnot{1}^T\binvar\Big) * \Big( \bm{B}\matnot{1}^T\bm{b}\matnot{1} \Big)$ and $\boutvar = \bm{C}\boutvar_{N} + \bm{\beta}$. The parameters $\bm{C} \in  \realnum^{o\times k}, \bm{A}\matnot{n} \in  \realnum^{d\times k}, \bm{S}\matnot{n} \in  \realnum^{k\times k}, \bm{B}\matnot{n} \in  \realnum^{\omega\times k}$, $\bm{b}\matnot{n} \in  \realnum^{\omega}$ for $n=1,\ldots,N$, are learnable.

\noindent\textbf{Model 3: \modelthree{} (Nested coupled CP decomposition with skip)}

The expressiveness of \modeltwo{} can be further extended using a skip connection (motivated by \modelone). The new model uses a nested coupled decomposition and has the following recursive expression:

\begin{equation}
    \boutvar_{n} = \Big(\bm{A}\matnot{n}^T\binvar\Big) * \Big(\bm{S}\matnot{n}^T \boutvar_{n-1} + \bm{B}\matnot{n}^T\bm{b}\matnot{n}\Big) +  \bm{V}\matnot{n}\boutvar_{n-1}
    \label{eq:prodpoly_model3}
\end{equation}

for $n=2,\ldots,N$ with $\boutvar_{1} = \Big(\bm{A}\matnot{1}^T\binvar\Big) * \Big( \bm{B}\matnot{1}^T\bm{b}\matnot{1} \Big)$ and $\boutvar = \bm{C}\boutvar_{N} + \bm{\beta}$. \rebuttal{The parameters $\bm{V}\matnot{n} \in  \realnum^{k\times k}$ are learnable}, while the rest parameters are the same as in \modeltwo. The difference in the recursive form results in a different polynomial expansion and thus architecture.

\begin{figure}[!h]
    \centering
    \includegraphics[width=1\linewidth]{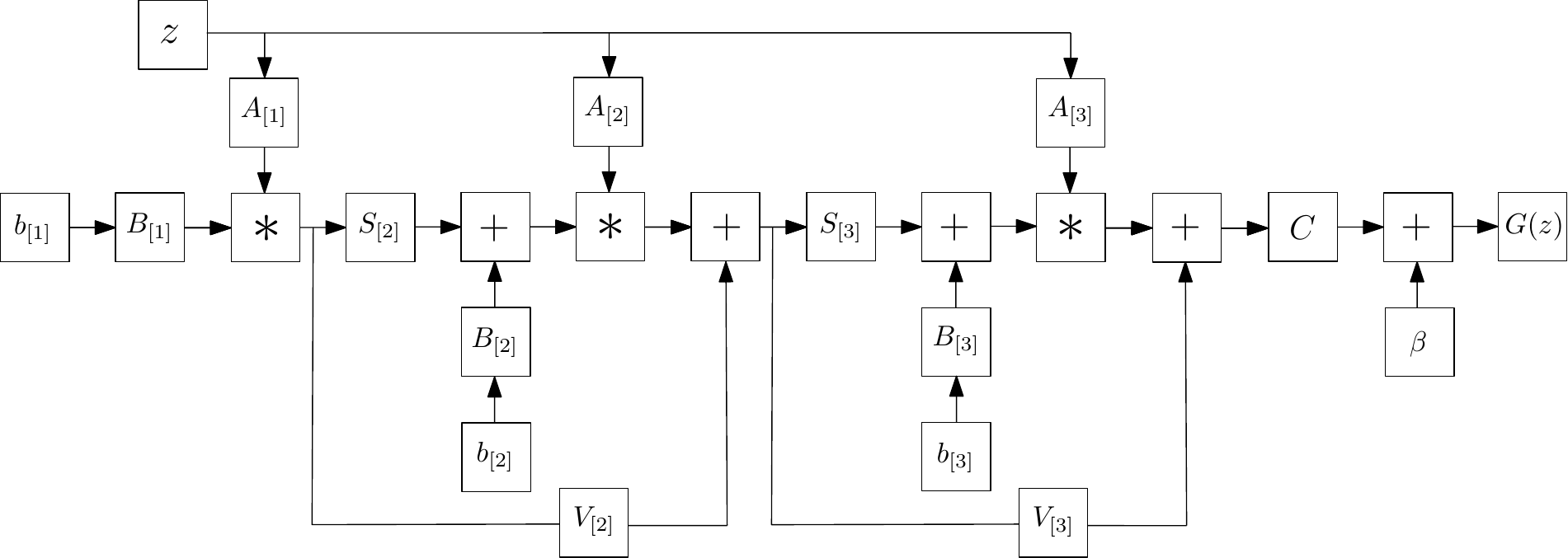}
\caption{Schematic illustration of the \modelthree{} (for third order approximation). The difference from Fig.~\ref{fig:prodpoly_model2_schematic} is the skip connections added in this model.}
\label{fig:prodpoly_model3_schematic}
\end{figure}

\noindent \textbf{Comparison between the models}

All three models (see Table~\ref{tbl:prodpoly_single_polynomial_models} for names and schematics) are based on a polynomial expansion, however their recursive forms and employed decompositions differ.

\modelone{} is a straightforward coupled decomposition and is a proof of concept that polynomials can learn high-dimensional distributions. \modeltwo{} illustrates how to convert a popular CNN/linear model of the form $\boutvar_{k} = \bm{S}\matnot{k}^T \boutvar_{k-1} + \bm{b}\matnot{k}$ to a polynomial (i.e., $\boutvar_{k} = (\bm{A}\matnot{k}^T\binvar) * (\bm{S}\matnot{k}^T \boutvar_{k-1} + \bm{b}\matnot{k})$). Similarly, \modelthree{} demonstrates how a residual network can be transformed into a polynomial. 

\rebuttal{An illustrative comparison of the three decompositions is conducted below. The challenging task of synthesizing images is selected. Each model is implemented using the respective decomposition, i.e., \modelone, \modeltwo, \modelthree. Following the derivations of the previous few paragraphs, we train a generator without activation functions between the blocks; a single hyperbolic tangent is used in the output space as a normalization\footnote{Further experiments without activation functions are deferred to the supplementary.}. The generator is trained with an adversarial loss, i.e., using Generative Adversarial Nets (GANs)~\cite{goodfellow2014generative}. GANs typically consist of two deep networks, namely a generator $G$ and a discriminator $D$. $G$ is a decoder, which receives as input a random noise vector $\bm{z} \in \realnum^{d}$ and outputs a sample $\bm{x} = G(\bm{z})$. $D$ receives as input both $G(\bm{z})$ and real samples and tries to differentiate the fake and the real samples. During training, both $G$ and $D$ compete against each other.}

\begin{figure*}[htb]
    \subfloat[GT]{\includegraphics[width=0.24\linewidth]{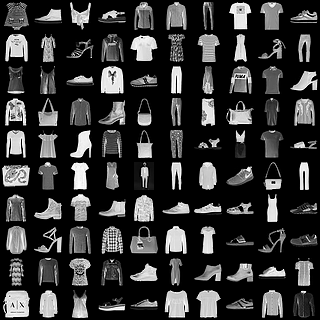}\hspace{1mm}}
    \subfloat[\modelname{} - \modelone]{\includegraphics[width=0.24\linewidth]{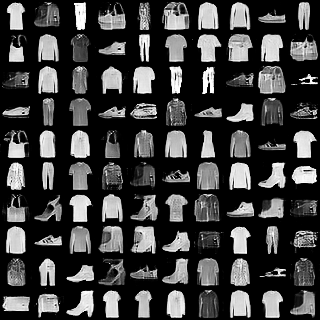}\hspace{1mm}}
    \subfloat[\modelname{} - \modeltwo]{\includegraphics[width=0.24\linewidth]{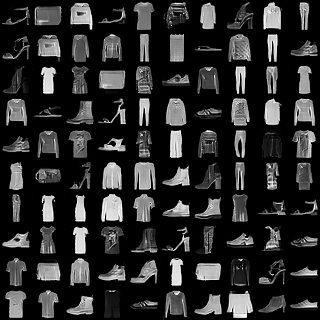}\hspace{1mm}}
    \subfloat[\modelname{} - \modelthree]{\includegraphics[width=0.24\linewidth]{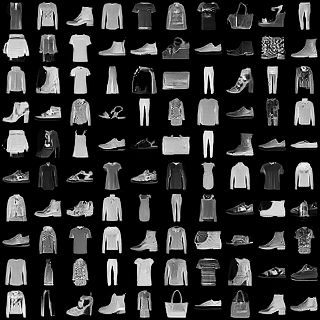}\hspace{1mm}}
\caption{Comparison of the proposed models in fashion image~\cite{xiao2017fashion} generation without activation functions.}
\label{fig:prodpoly_linear_fashionmnist_model_comparison}
\end{figure*}

\rebuttal{The three models are originally compared in fashion image generation. The outcomes, visualized in Fig.~\ref{fig:prodpoly_linear_fashionmnist_model_comparison}, demonstrate similar generation properties.}

\begin{figure*}[htb]
    \subfloat[GT]{\includegraphics[width=0.24\linewidth]{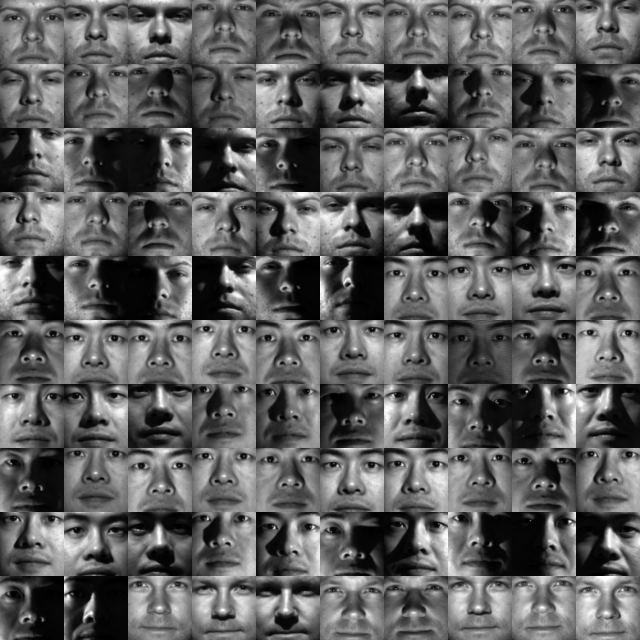}\hspace{1mm}}
    \subfloat[\modelname{} - \modelone]{\includegraphics[width=0.24\linewidth]{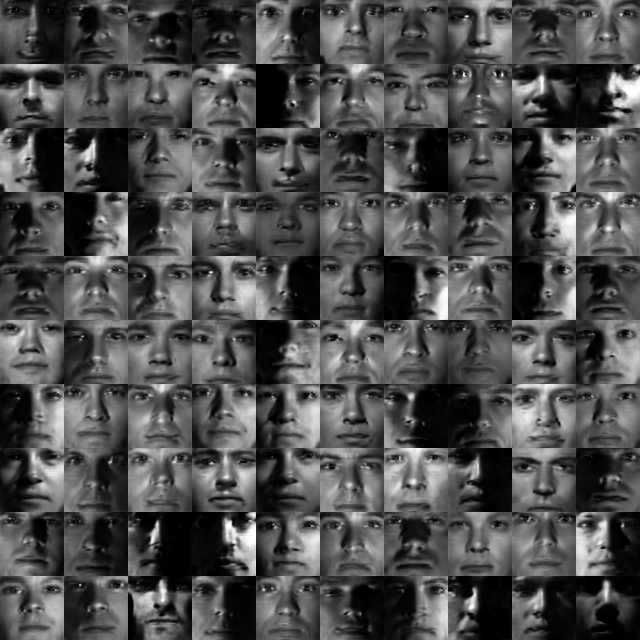}\hspace{1mm}}
    \subfloat[\modelname{} - \modeltwo]{\includegraphics[width=0.24\linewidth]{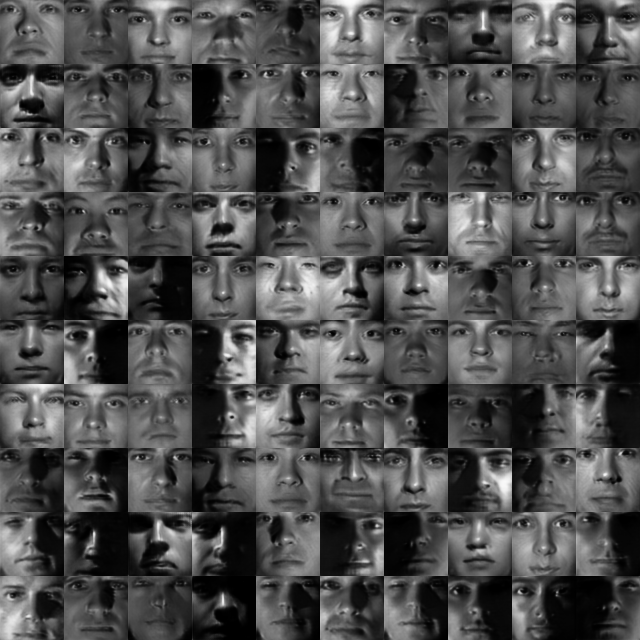}\hspace{1mm}}
    \subfloat[\modelname{} - \modelthree]{\includegraphics[width=0.24\linewidth]{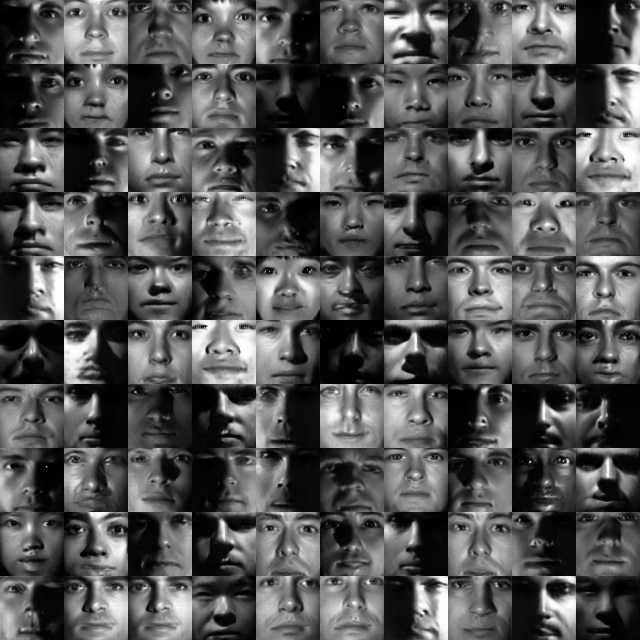}\hspace{1mm}}
\caption{Comparison of the proposed models in facial image~\cite{georghiades2001few} generation without activation functions.}
\label{fig:prodpoly_linear_yaleb_model_comparison}
\end{figure*}

\rebuttal{In Fig.~\ref{fig:prodpoly_linear_yaleb_model_comparison}, samples of the three models are synthesized when trained facial images. All three models can generate faces without activation functions between the layers, while the three models share similar generation quality.}

In the remainder of the paper, for comparison purposes we use the \modeltwo{} by default for the image generation and \modelthree{} for the image classification. In all cases, to mitigate stability issues that might emerge during training, we employ certain normalization schemes that constrain the magnitude of the gradients.

\subsection{Product of polynomials}
\label{ssec:prodpoly_product_poly}
\begin{figure}[!h]
    \centering
    \includegraphics[width=1\linewidth]{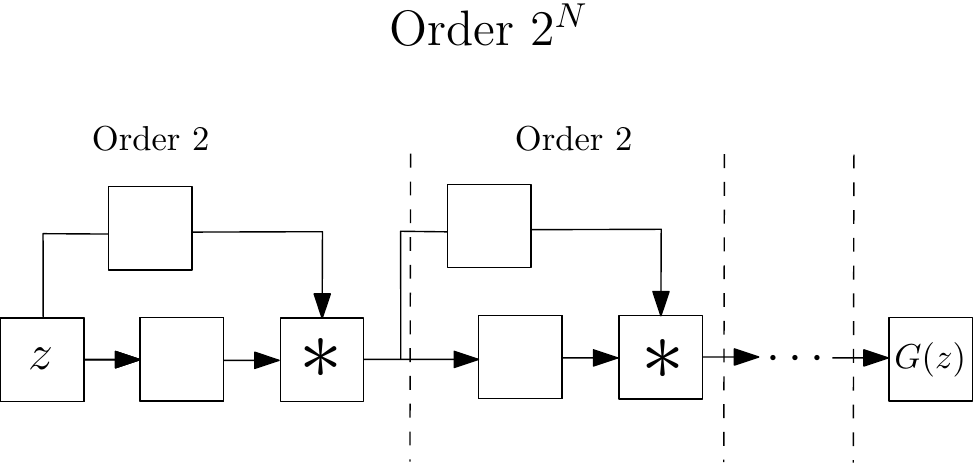}
\caption{Abstract illustration of the \modelname. The input variable $\binvar$ on the left is the input to a $2^{nd}$ order expansion; the output of this is used as the input for the next polynomial (also with a $2^{nd}$ order expansion) and so on. If we use $N$ such polynomials, the final output $G(\binvar)$ expresses a $2^N$ order expansion. In addition to the high order of approximation, the benefit of using the product of polynomials is that the model is flexible, in the sense that each polynomial can be implemented as a different decomposition of Sec.~\ref{ssec:prodpoly_single_poly}.}
\label{fig:prodpoly_prod_schematic}
\end{figure}

Instead of using a single polynomial, we express the function approximation as a product of polynomials. The product is implemented as successive polynomials where the output of the $i^{th}$ polynomial is used as the input for the $(i+1)^{th}$ polynomial. The concept is visually depicted in Fig.~\ref{fig:prodpoly_prod_schematic}; each polynomial expresses a second order expansion. Stacking $N$ such polynomials results in an overall order of $2^N$. Trivially, if the approximation of each polynomial is $B$ and we stack $N$ such polynomials, the total order is $B^N$. The product does not necessarily demand the same order in each polynomial, the model and the expansion order of each polynomial can be different and dependent on the task. For instance, for generative tasks that the resolution increases progressively, the expansion order could increase in the last polynomials. In all cases, the final order will be the product of each polynomial power.

There are two main benefits of the product over the single polynomial: a) it allows using different decompositions (e.g., like in Sec.~\ref{ssec:prodpoly_single_poly}) and expansion order for each polynomial; b) it requires much fewer parameters for achieving the same order of approximation. Given the benefits of the product of polynomials, we assume below that a product of polynomials is used, unless explicitly mentioned otherwise. The respective model of product polynomials is called \modelname.

\subsection{Task-dependent input/output}
\label{ssec:prodpoly_method_details}

The aforementioned polynomials are a function $\boutvar = G(\binvar)$, where the input/output are task-dependent. For a generative task, e.g., learning a decoder, the input $\binvar$ is typically some low-dimensional noise, while the output is a high-dimensional signal, e.g., an image. For a discriminative task the input $\binvar$ is an image; for a domain adaptation task the signal $\binvar$ denotes the source domain and $\boutvar$ the target domain.

 \section{Experiments}
\label{sec:prodpoly_experiments}
We conduct four experiments against state-of-the-art models in three diverse tasks\footnote{The source code is available in \url{https://github.com/grigorisg9gr/polynomial_nets}.}: image generation, image classification, face verification/identification and graph representation learning. In each case, the baseline considered is converted into an instance of our family of $\Pi$-nets and the two models are compared. \rebuttal{Experiments on image generation and image classification without using activation functions between the layers are deferred to the supplementary.}

\subsection{Image generation}
\label{ssec:prodpoly_experiment_stylegan}
The robustness of \modelname{} in image generation is assessed in two different architectures/datasets below. 

\textbf{SNGAN on CIFAR10}: In the first experiment, the architecture of SNGAN~\cite{miyato2018spectral} is selected as a strong baseline on CIFAR10~\cite{krizhevsky2014cifar}. The baseline includes $3$ residual blocks in the generator and the discriminator. 

The generator is converted into a $\Pi$-net, where each residual block is a single order of the polynomial. We implement two versions, one with a single polynomial (\modeltwo) and one with product of polynomials (where each polynomial uses \modeltwo). In our implementation $\bm{A}_{[n]}$ is a thin FC layer, $(\bm{B}\matnot{n})^T \bm{b}\matnot{n}$ is a bias vector and $\bm{S}\matnot{n}$ is the transformation of the residual block. Other than the aforementioned modifications, the hyper-parameters (e.g., discriminator, learning rate, optimization details) are kept the same as in SNGAN~\cite{miyato2018spectral}. 

Each network was run for 10 times and the mean and variance are reported. The popular Inception Score (IS)~\citep{salimans2016improved} and the Frechet Inception Distance (FID)~\citep{heusel2017gans} are used for quantitative evaluation. Both scores extract feature representations from a pre-trained classifier (the Inception network~\citep{szegedy2015going}). 

The quantitative results are summarized in Table~\ref{tab:prodpoly_exper_cifar10_sota}. In addition to SNGAN and our two variations with polynomials, we have added the scores of \cite{grinblat2017class,gulrajani2017improved,du2019implicit, hoshen2019non,lucas2019adversarial} as reported in the respective papers. Note that the single polynomial already outperforms the baseline, while the \modelname{} boosts the performance further and achieves a substantial improvement over the original SNGAN.

\begin{table}
    \caption{IS/FID scores on CIFAR10~\citep{krizhevsky2014cifar} generation. The scores of \cite{grinblat2017class,gulrajani2017improved} are added from the respective papers as using similar residual based generators. The scores of \cite{du2019implicit, hoshen2019non, lucas2019adversarial} represent alternative generative models. \modelname\ outperforms the compared methods in both metrics.}
     \begin{tabular}{|c | c | c|} 
     \hline
     \multicolumn{3}{|c|}{Image generation on CIFAR10}\\ 
     \hline
     Model & IS ($\uparrow$) & FID ($\downarrow$)\\
     \hline
     SNGAN & $8.06\pm 0.10$ & $19.06\pm 0.50$\\
     \hline
     \modeltwo (Sec.~\ref{ssec:prodpoly_single_poly}) & $8.30\pm 0.09$ & $17.65\pm 0.76$\\
     \hline
     \modelname & $\bm{8.49\pm 0.11}$ & $\bm{16.79\pm 0.81}$\\
     \cmidrule[\heavyrulewidth](){1 - 3}
     CSGAN-\cite{grinblat2017class} & $7.90\pm0.09$ & -\\
     \hline
     WGAN-GP-\cite{gulrajani2017improved} & $7.86\pm0.08$ & -\\
     \hline
     CQFG-\cite{lucas2019adversarial}  & $8.10$ & $18.60$\\
     \hline
     EBM~\cite{du2019implicit} & $6.78$ & $38.2$ \\
     \hline
     GLANN~\cite{hoshen2019non} & - & $46.5\pm0.20$\\
      \hline
     \end{tabular}
     \label{tab:prodpoly_exper_cifar10_sota}
\end{table}

\textbf{StyleGAN on FFHQ}: StyleGAN~\cite{karras2018style} is the state-of-the-art architecture in image generation. The generator is composed of two parts, namely: (a) the mapping network, composed of 8 FC layers, and (b) the synthesis network, which is based on ProGAN~\cite{karras2017progressive} and progressively learns to synthesize high quality images. The sampled noise is transformed by the mapping network and the resulting vector is then used for the synthesis network. As discussed in the introduction, StyleGAN is already an instance of the $\Pi$-net family, due to AdaIN. 
Specifically, the $k^{th}$ AdaIN layer is $\bm{h}_k = (\bm{A}_k^T\bm{w}) * n(c(\bm{h}_{k-1}))$, where $n$ is a normalization, $c$ the convolution operator and $\bm{w}$ is the transformed noise $\bm{w} = MLP(\bm{z})$ (mapping network). 
This is equivalent to our NCP model by setting $\bm{S}\matnot{k}^T$ as the convolution operator. 

In this experiment we illustrate how simple modifications, using our family of products of polynomials, further improve the representation power. We make a minimal modification in the mapping network, while fixing the rest of the hyper-parameters. In particular, we convert the mapping network into a polynomial (specifically a \modeltwo), which makes the generator a product of two polynomials.  

\begin{figure*}
\centering
    \centering
    \includegraphics[width=0.123\linewidth]{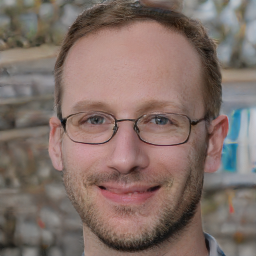}\hspace{-0.5mm}
    \includegraphics[width=0.123\linewidth]{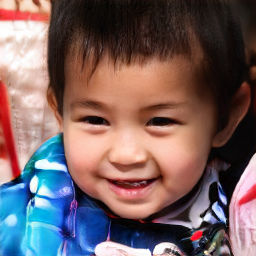}\hspace{-0.5mm}
    \includegraphics[width=0.123\linewidth]{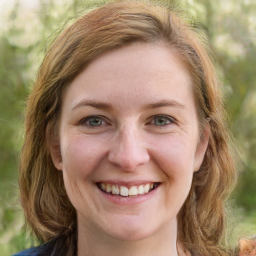}\hspace{-0.5mm}
    \includegraphics[width=0.123\linewidth]{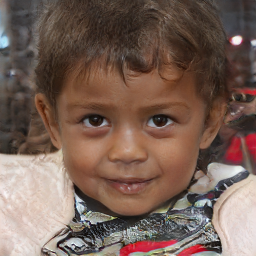}\hspace{-0.5mm}
    \includegraphics[width=0.123\linewidth]{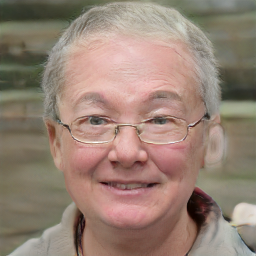}\hspace{-0.5mm}
    \includegraphics[width=0.123\linewidth]{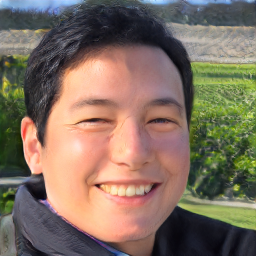}\hspace{-0.5mm}
    \includegraphics[width=0.123\linewidth]{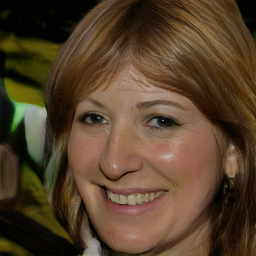}\hspace{-0.5mm}
    \includegraphics[width=0.123\linewidth]{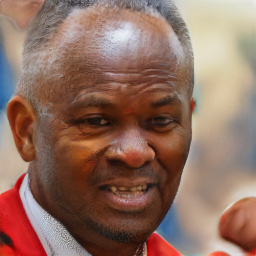} 
\caption{Samples synthesized from \modelname{} (trained on FFHQ).}
\label{fig:prodpoly_stylegan_visual}
\end{figure*}

The Flickr-Faces-HQ Dataset (FFHQ) dataset~\cite{karras2018style} which includes $70,000$ images of high-resolution faces is used. All the images are resized to $256\times256$. The best FID scores of the two methods (in $256\times 256$ resolution) are $\bm{6.82}$ for ours and $7.15$ for the original StyleGAN, respectively. That is, our method improves the results by $5\%$. Synthesized samples of our approach are visualized in Fig.~\ref{fig:prodpoly_stylegan_visual}.

\subsection{Classification}
We perform two experiments on classification: a) audio classification, b) image classification. Residual Network (\resnet)~\cite{he2016deep,srivastava2015highway} and its variants~\cite{huang2017densely, wang2018mixed, xie2017aggregated, zhang2017residual, zagoruyko2016wide} have been applied to diverse tasks including object detection and image generation~\cite{grinblat2017class,gulrajani2017improved,miyato2018spectral}. 
The core component of \resnet{} is the residual block; the $t^{th}$ residual block is expressed as $\bm{z}_{t+1} = \bm{z}_t + \bm{C} \bm{z}_t$ for input $\bm{z}_t$. Each residual block is adapted (using \modelthree) to express a higher-order expansion; \rebuttal{this is achieved by using $\bm{V}\matnot{n} = \bm{I} + \bm{S}\matnot{n}$ in \eqref{eq:prodpoly_model3}, where $\bm{I}$ is an identity matrix.} The output of each residual block is the input for the next residual block, which makes our \resnet{} a product of polynomials.

\textbf{Audio classification}: \rebuttal{The goal of this experiment is to reduce the number of residual blocks (of higher-order polynomial expansion) without sacrificing the performance of the original \resnet, while validating the performance of the proposed method in a distribution that differs from that of natural images.}

The performance of \resnet{} is evaluated on the Speech Commands dataset~\cite{warden2018speech}. 
The dataset includes $60,000$ audio files; each audio contains a single word of a duration of one second. There are $35$ different words (classes) with each word having $1,500 - 4,100$ recordings. Every audio file is converted into a mel-spectrogram of resolution $32\times32$. Each method is trained for $70$ epochs with SGD and initial learning rate of $0.01$. The learning rate is reduced if the validation accuracy does not improve for two consecutive epochs.

The baseline is a \resnet34 architecture; we use second-order residual blocks to build the \modelres{} to match the performance of the baseline.
The quantitative results are added in Table~\ref{tab:prodpoly_resnet_speech_command}. The two models share the same accuracy, however \modelres{} includes $38\%$ fewer parameters. \rebuttal{This result validates our assumption that $\Pi$-nets can achieve the same performance with less parameters than the baseline.}

\begin{table}[h]
 \caption{Speech classification with \resnet. The accuracy of the compared methods is similar, but \modelres{} has $38\%$ fewer parameters. The symbol `\# par' abbreviates the number of parameters (in millions). }
\centering
     \begin{tabular}{|c | c | c | c|} 
         \hline
         \multicolumn{4}{|c|}{Speech Commands classification with \resnet}\\ 
         \hline
         Model & \# blocks & \# par & Accuracy\\
        \hline
         \resnet34 & $[3, 4, 6, 3]$ &  $21.3$ & $0.951 \pm 0.002$\\
         \hline
         \modelres & $[3, 3, 3, 2]$ &  $\bm{13.2}$ & $0.951 \pm 0.002$\\
         \hline
     \end{tabular}
 \label{tab:prodpoly_resnet_speech_command}
\end{table}

\textbf{Classification on CIFAR}: \rebuttal{The performance of the polynomial networks is also assessed on CIFAR10 and CIFAR100 classification. The goal is to reduce the number of residual blocks (of higher-order polynomial expansion) without sacrificing the performance of the original \resnet.}

We select the \resnet18 and \resnet34 as baselines. 
Each method is trained for $120$ epochs with batch size $128$. The SGD optimizer is used with initial learning rate of $0.1$. The learning rate is multiplied with a factor of $0.1$ in epochs $40, 60, 80, 100$.

\begin{table}[h]
 \caption{Image classification on CIFAR10 with \resnet. The \# abbreviates `number of', while the parameters are measured in millions. The term `block' abbreviates a `residual block'. Note that each baseline, e.g. \resnet18, has the same performance with the respective \modelres, but significantly more parameters.}
\centering
     \begin{tabular}{|c | c | c | c|} 
         \hline
         \multicolumn{4}{|c|}{CIFAR10 classification with \resnet}\\ 
         \hline
         Model & \# blocks & \# params (M) & Accuracy\\
        \hline
         \resnet18 & $[2, 2, 2, 2]$ &  $11.2$ & $0.945 \pm 0.000$\\
         \hline
         \modelres & $[2, 2, 1, 1]$ &  $\bm{6.0}$ & $0.945 \pm 0.001$\\
         \hline\hline
         \resnet34 & $[3, 4, 6, 3]$ &  $21.3$ & $0.948 \pm 0.001$\\
         \hline
         \modelres & $[3, 3, 2, 2]$ &  $\bm{13.0}$ & $0.949 \pm 0.002$\\
         \hline
     \end{tabular}
 \label{tab:prodpoly_resnet_cifar10}
\end{table}

\begin{figure}[h]
    \centering
    \subfloat[\resnet18]{\includegraphics[width=0.48\linewidth]{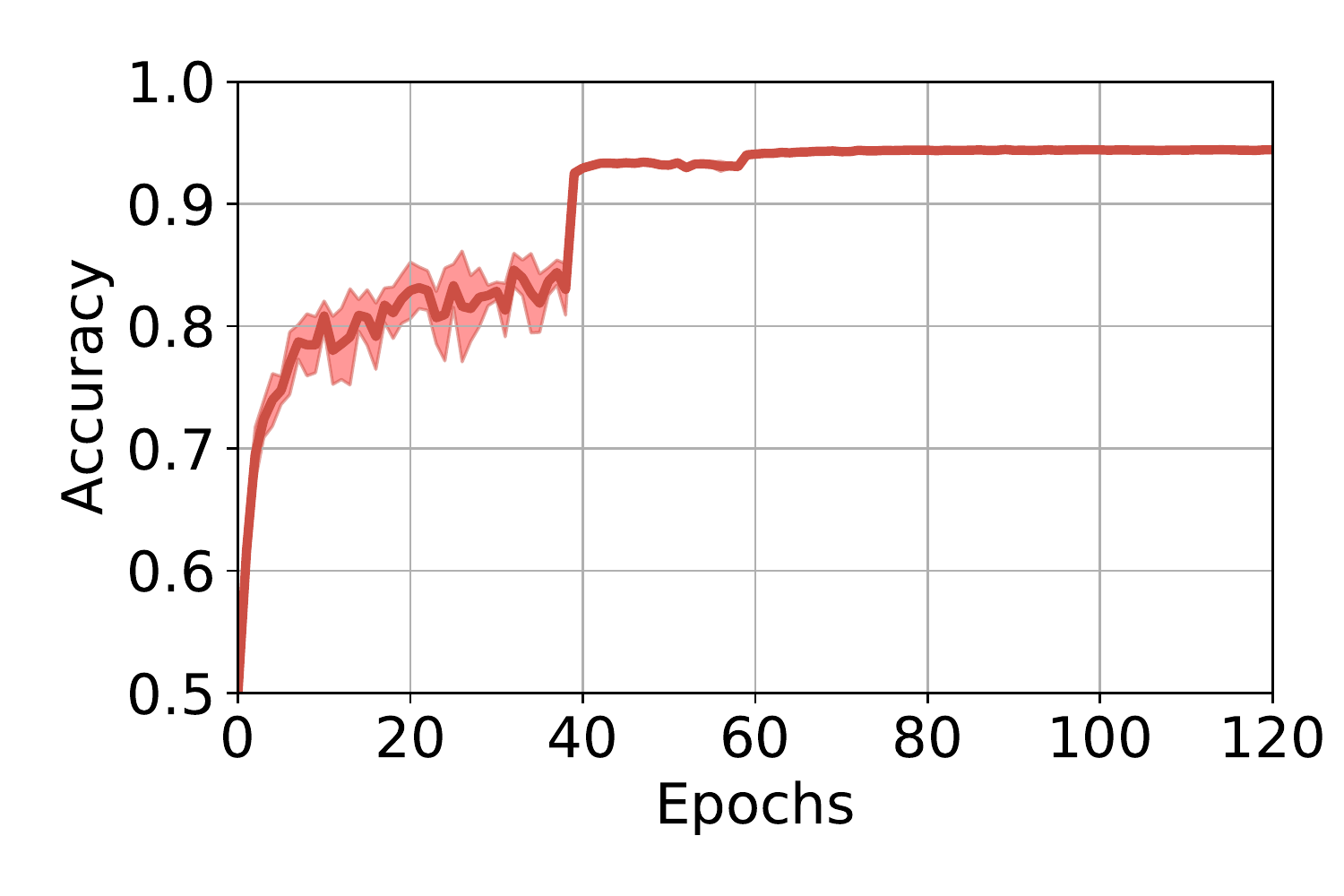}\hspace{-0.1mm}}
  \quad
    \subfloat[\modelres]{\includegraphics[width=0.48\linewidth]{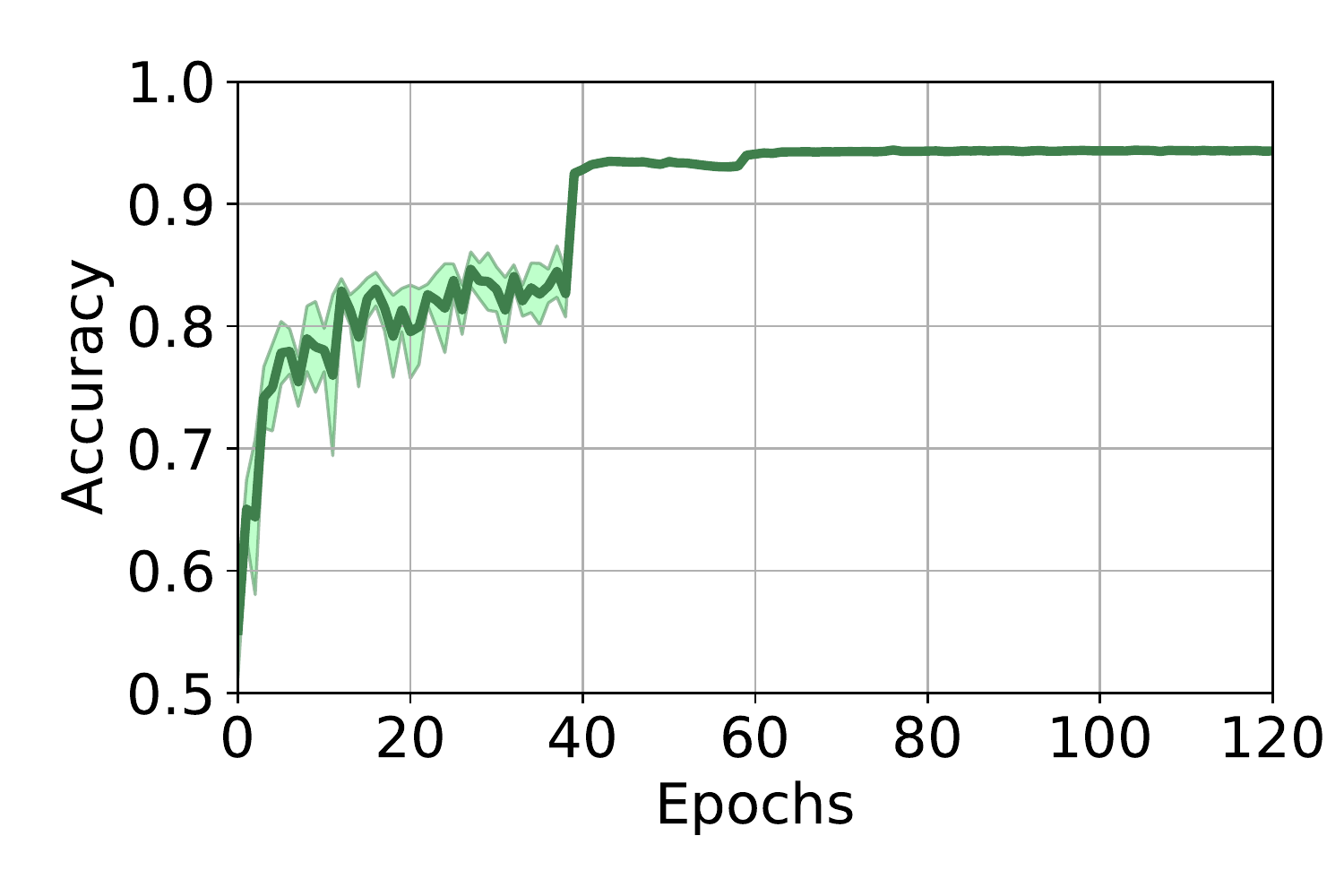}\hspace{-1.9mm}}
\caption{The test accuracy of (a) \resnet18 and (b) the respective \modelres{} are plotted (CIFAR10 training). The two models perform similarly throughout the training, while ours has $46\%$ less parameters. The width of the highlighted region denotes the standard deviation of each model.}
\label{fig:prodpoly_resnet_accuracy_bars_resnet18}
\end{figure}

In Table~\ref{tab:prodpoly_resnet_cifar10} the two different \resnet{} baselines are compared against \modelres\ on CIFAR10; the respective \modelres{} models have the same accuracy. However, each \modelres{} has $\sim 40\%$ less parameters than the respective baseline. In addition, we visualize the test accuracy for \resnet18 and the respective \modelres{} in Fig.~\ref{fig:prodpoly_resnet_accuracy_bars_resnet18}. The test error of the two models is similar throughout the training. 

The same experiment is repeated on CIFAR100 with \resnet34 as the baseline. Table~\ref{tab:prodpoly_resnet_cifar100} exhibits a similar pattern. That is, the test accuracy of \resnet34 and \modelres{} is similar, however \modelres{} has $\sim 30\%$ less parameters.

\begin{table}[h]
 \caption{CIFAR100 classification with \resnet. The accuracy of the compared methods is similar, but \modelres{} has $30\%$ less parameters.}
\centering
     \begin{tabular}{|c | c | c | c|} 
         \hline
         \multicolumn{4}{|c|}{CIFAR100 classification with \resnet}\\ 
         \hline
         Model & \# blocks & \# params (M) & Accuracy\\
        \hline
         \resnet34 & $[3, 4, 6, 3]$ &  $21.3$ & $0.769 \pm 0.003$\\
         \hline
         \modelres & $[3, 4, 3, 2]$ &  $\bm{14.7}$ & $0.769 \pm 0.001$\\
         \hline
     \end{tabular}
 \label{tab:prodpoly_resnet_cifar100}
\end{table}

\textbf{Classification on ImageNet}:
We perform a large-scale classification experiment on ImageNet~\cite{russakovsky2015imagenet}. Models are trained on a DGX station with 4 Tesla V100 (32GB) GPUs. 
To stabilize the training, the second order of each residual block is normalized with a hyperbolic tangent unit. SGD with momentum $0.9$, weight decay $10^{-4}$ and a mini-batch size of $512$ is used. The initial learning rate is set to $0.2$ and decreased by a factor of $10$ at $30, 60$, and $80$ epochs. Models are trained for $90$ epochs from scratch, using linear warm-up of the learning rate during first five epochs according to \citet{goyal2017accurate}. 

\begin{figure*}[h!]
\centering
\subfloat[Top-1 Error]{
\label{pic:imagenettop1} 
\includegraphics[width=0.5\textwidth]{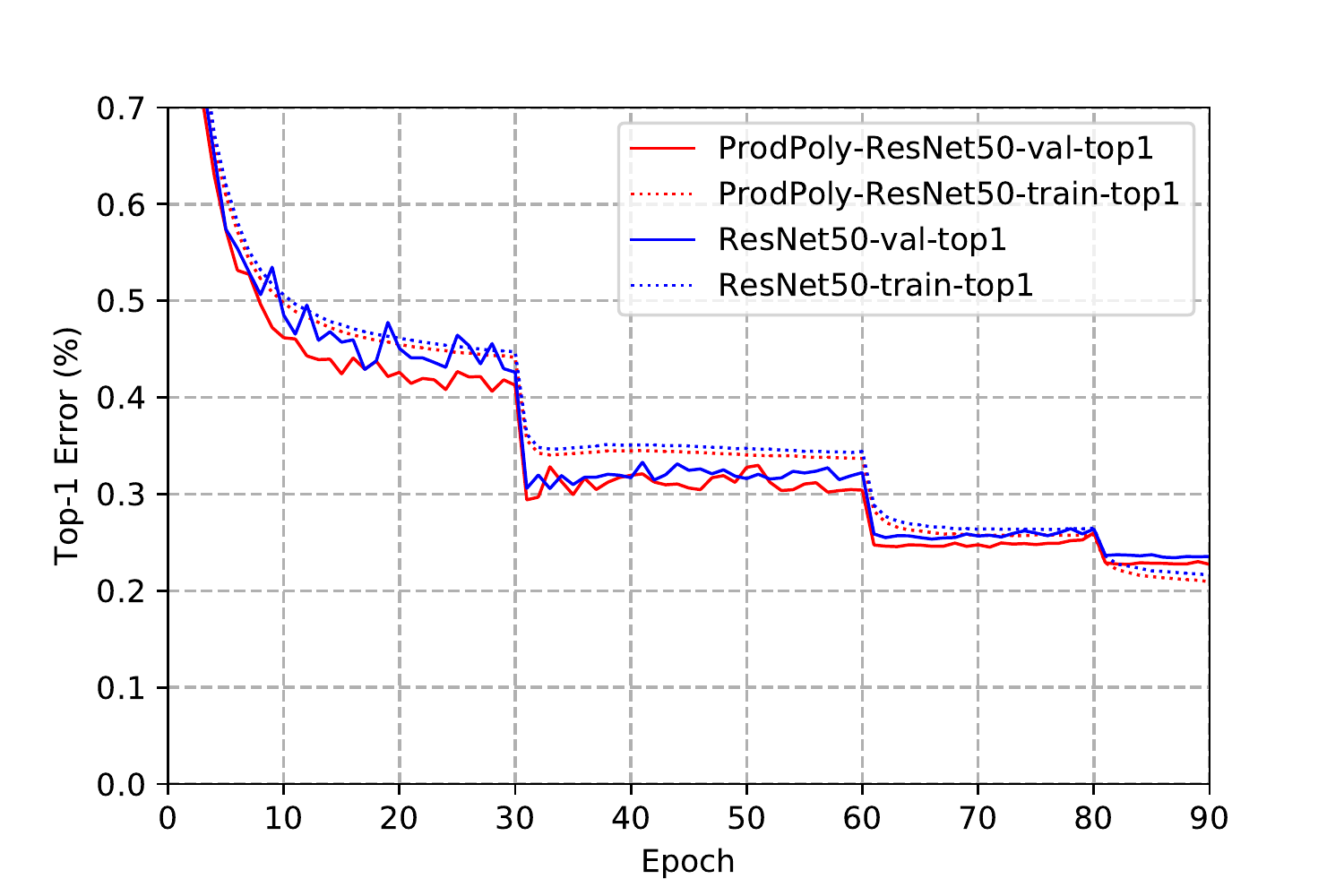}}
\subfloat[Top-5 Error]{
\label{pic:imagenettop5} 
\includegraphics[width=0.5\textwidth]{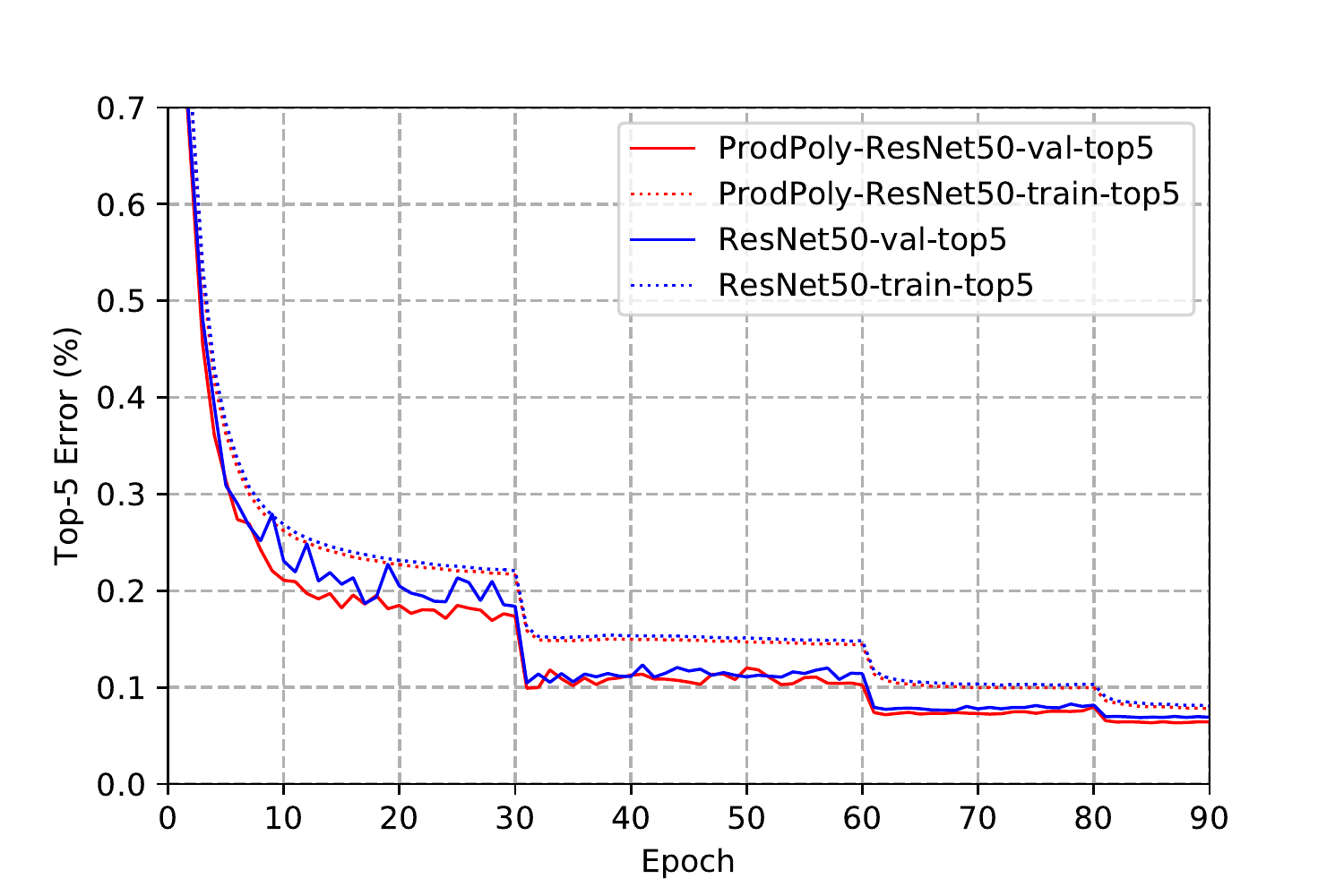}}
\caption{Top-1 and Top-5 error curves on the ImageNet dataset.}
\label{pic:imagenet}
\end{figure*}

The Top-1 and Top-5 error throughout the training is visualized in Fig.~\ref{pic:imagenet}, while the validation results are added in Table~\ref{tab:prodpoly_resnet_imagenetfp32}. For a fair comparison, we report the results from our training in both the original \resnet{} and \modelres{}\footnote{The performance of the original \resnet~\citet{he2016deep} is inferior to the one reported here and in \citet{hu2018squeeze}.}. \modelres{} consistently improves the performance with a negligible increase in computational complexity. Remarkably, \modelres50 achieves a single-crop Top-1 validation error of $22.827\%$ and Top-5 validation error of $6.431\%$, exceeding \resnet50 by $0.719\%$ and $0.473\%$, respectively.

\begin{table}[h!]
\caption{ImageNet classification results of ResNet50 and the proposed \modelres50. ``Throughput'' denotes the total Images Per Second (IPS) during training.}
\label{tab:prodpoly_resnet_imagenetfp32}
\centering
\begin{tabular}{c | c | c | c}
\hline
Model & Top-1 error ($\%$) & Top-5 error ($\%$) & Throughput\\
\hline
\resnet50   & 23.546  & 6.904                   & 1625 \\
\modelres50 & 22.827  \tiny{($\downarrow 0.719$)}  & 6.431 \tiny{($\downarrow 0.473$)} & 1531\\
\hline
\end{tabular}
\end{table}

\subsection{Face verification and identification}
We scrutinize the performance of the $\Pi$-nets on the challenging task of face recognition. The architecture of the current state-of-the-art method of ArcFace~\cite{deng2019arcface} is a \resnet, which we can convert into a polynomial network using the \modelthree. 

\noindent{\bf Training Data}: The data of MS1M-RetinaFace dataset~\cite{guo2016ms,deng2019lightweight} consist the training images; all face images inside MS1M-RetinaFace are pre-processed to the size of $112\times112$ based on the five facial landmarks predicted by RetinaFace~\cite{deng2019retinaface}. In total, there are 5.1M images of 93K identities. 

\noindent{\bf Testing Data}: The performance is compared on widely used face verification data-sets (e.g., LFW~\cite{huang2008labeled}, CFP~\cite{sengupta2016frontal}, AgeDB~\cite{Moschoglou2017AgeDB}, CPLFW~\cite{zheng2018cross}, CALFW~\cite{zheng2017cross} and RFW~\cite{Wang2019racial}).
Besides, we also extensively test the proposed method on large-scale benchmarks (e.g., IJB-B \cite{whitelam2017iarpa}, IJB-C \cite{maze2018iarpa} and MegaFace \cite{kemelmacher2016megaface}); the fundamental statistics of all the datasets are summarized in Table~\ref{table:dataset}. To get the embedding features for templates (e.g., IJB-B \cite{whitelam2017iarpa} and IJB-C \cite{maze2018iarpa}), we simply calculate the feature center of all images from the template or all frames from the video. 

\noindent{\bf Training Details}: For the baseline embedding network, we employ the widely used CNN architecture, \resnet50. 
Specifically, we follow \cite{deng2019arcface} to set the feature scale $s$ to 64 and choose the angular margin $m$ at $0.5$. 
The batch size is set to $512$ with momentum $0.9$ and weight decay $5e-4$, while we decrease the learning rate in iterations 100K, 160K, 220K. The training finishes after 30 epochs. The implementation is by MXNet \cite{chen2015mxnet,deng2019arcface} and the models are trained on 8 NVIDIA 2080ti (11GB) GPUs. 

The baseline residual block is converted into a second-order residual block to build the \modelres{}, while we keep all the other settings exactly the same as the baseline. After training, we only keep the feature embedding network without the fully connected layer (174.5MB for \resnet50 and 181.8MB for \modelres50) and extract the $512$-$D$ features ($5.76$ ms/face for \resnet50 and $6.02$ ms/face for \modelres50) for each normalised face. Compared to \resnet50, \modelres50 obviously boosts the performance only by a negligible increase in model size and latency.

\begin{table}[t!]
\begin{center}
\caption{Face datasets for training and testing. ``(P)'' and ``(G)'' refer to the probe and gallery set, respectively.}
\label{table:dataset}
\begin{tabular}{c|c|c}
\hline
 Datasets   & \#Identity & \#Image\\
\hline
MS1MV2                             & 93K  & 5.1M \\   
\hline
LFW \cite{huang2008labeled}   & 5,749  & 13,233 \\
CFP \cite{sengupta2016frontal} & 500 & 7,000\\
AgeDB \cite{Moschoglou2017AgeDB} & 568 & 16,488 \\
CPLFW \cite{zheng2018cross} & 5,749  & 11,652 \\
CALFW \cite{zheng2017cross} & 5,749  & 12,174 \\
RFW   \cite{Wang2019racial} & 11,430  & 40,607 \\ 
RFW-Caucasian \cite{Wang2019racial} & 2,959 & 10,196\\
RFW-Indian \cite{Wang2019racial} & 2,984 & 10,308\\
RFW-Asian  \cite{Wang2019racial} & 2,492 & 9,688 \\
RFW-African \cite{Wang2019racial} & 2,995 & 10,415\\
\hline
MegaFace \cite{kemelmacher2016megaface} & 530 (P) & 1M (G) \\
IJB-B    \cite{whitelam2017iarpa}       & 1,845   & 76.8K \\
IJB-C    \cite{maze2018iarpa}           & 3,531   & 148.8K \\
\hline
\end{tabular}
\end{center}
\end{table}

\noindent {\bf Results on LFW, CFP-FF, CFP-FP, CPLFW, AgeDB-30, CALFW and RFW.} 
LFW \cite{huang2008labeled} contains 13,233 web-collected images from 5,749 different identities, with limited variations in pose, age, expression and illuminations.
CFP \cite{sengupta2016frontal} consists of collected images of celebrities in frontal and profile views. 
On CFP, there are two evaluation protocols: CFP-Frontal-Frontal and CFP-Frontal-Profile. CFP-Frontal-Profile is very challenging as the pose gap within positive pairs is around $90^{\circ}$.
CPLFW \cite{zheng2018cross} was collected by crowd-sourcing efforts to seek the pictures of people in LFW with pose gap as large as possible from the Internet. 
CALFW \cite{zheng2017cross} is similar to CALFW, but from the perspective of age difference.
AgeDB \cite{Moschoglou2017AgeDB} contains manually annotated images. In this paper, we use the evaluation protocol with 30 years gap \cite{deng2019arcface}. 
RFW \cite{Wang2019racial} is a benchmark for measuring racial bias, which consists of four test subsets, namely Caucasian, Indian, Asian and
African. 
The quantitative results of the comparisons are exhibited in Table \ref{table:verification11}.
On LFW and CFP-FP, the results of \resnet50 and \modelres50 are similar to face verification on semi-frontal faces is saturated.
Nevertheless, \modelres50 significantly outperforms \resnet50 on CFP-FP, CPLFW, AgeDB-30, CALFW and RFW, indicating that the proposed method can enhance the robustness of the embedding features under pose variations, age variations and racial variations. 

\begin{table}[t!]
\begin{center}
\caption{Verification performance ($\%$) of \resnet50 and the proposed \modelres50 on LFW, CFP-FF, CFP-FP, CPLFW, AgeDB-30, CALFW and RFW (Caucasian, Indian, Asian and African).}
\label{table:verification11}
\begin{tabular}{c|c|c}
\hline
Method & \resnet50 & \modelres50 \\
\hline
LFW    & $ 99.733\pm0.309$ & {\bf 99.833$\pm0.211$} \tiny{($\uparrow 0.100$)} \\
CFP-FF & $ 99.871\pm0.135$ & {\bf 99.886$\pm0.178$} \tiny{($\uparrow 0.015$)}\\
\hline
CFP-FP & $ 98.800\pm0.249$ & {\bf 98.986$\pm0.274$} \tiny{($\uparrow 0.186$)} \\
CPLFW & $ 92.433\pm1.245$ & {\bf 93.317$\pm1.343$} \tiny{($\uparrow 0.884$)} \\ 
\hline
AgeDB-30 & $ 98.233\pm0.655$ & {\bf 98.467$\pm0.623$} \tiny{($\uparrow 0.234$)}\\
CALFW & $ 95.917\pm1.209$ & {\bf 96.233$\pm1.114 $} \tiny{($\uparrow 0.316$)}\\
\hline
RFW-Caucasian & $ 99.333\pm0.307$ & {\bf 99.700$\pm0.100 $} \tiny{($\uparrow 0.367$)}\\
RFW-Indian  & $ 98.567\pm0.507$ & {\bf 99.300$\pm0.296 $} \tiny{($\uparrow 0.733$)}\\
RFW-Asian  & $ 98.333\pm0.435$ & {\bf 98.950$\pm0.350 $} \tiny{($\uparrow 0.617$)}\\
RFW-African & $ 98.650\pm0.329$ & {\bf 99.417$\pm0.227 $} \tiny{($\uparrow 0.767$)}\\
\hline
\end{tabular}
\end{center}
\end{table}

\begin{table*}[t!]
\begin{center}
\caption{\textbf{$1$:$1$ verification TAR} on the IJB-B and IJB-C datasets.}
\label{tab:comp_ijb1vs1}
\begin{tabular}{c|cccc|cccc}
\hline
\multirow{2}{*}{Methods  (\%) }        &  \multicolumn{4}{c}{IJB-B} & \multicolumn{4}{c}{IJB-C}\\ 
                         \cmidrule(r){2-5} \cmidrule{6-9}
                        &FAR=$1e$$-$$6$ &FAR=$1e$$-$$5$ & FAR=$1e$$-$$4$ & FAR=$1e$$-$$3$ 
                        &FAR=$1e$$-$$6$ &FAR=$1e$$-$$5$ & FAR=$1e$$-$$4$ & FAR=$1e$$-$$3$ \\
                         \hline
ResNet50   & 37.28 & 90.73 & 94.73  & 96.63
           & 90.47 & 94.28 & 96.17  & 97.57 \\
Prodpoly-ResNet50  & {\bf 43.46} \tiny{($\uparrow 6.18$)} & {\bf 91.95} \tiny{($\uparrow 1.22$)} & {\bf 95.19} \tiny{($\uparrow 0.46$)} & {\bf 96.67} \tiny{($\uparrow 0.04$)}
                   & {\bf 90.77} \tiny{($\uparrow 0.30$)}& {\bf 95.16} \tiny{($\uparrow 0.88$)}& {\bf 96.58} \tiny{($\uparrow 0.41$)} & {\bf 97.66} \tiny{($\uparrow 0.09$)} \\\hline
\end{tabular}
\end{center}
\end{table*}

\begin{table*}[t!]
\begin{center}
\caption{\textbf{1:N (mixed media) Identification} on the IJB-B and IJB-C datasets. False positive identification rate (FPIR) is the proportion of non-mated searches returning any (1 or more) candidates at or above a threshold.}
\label{tab:comp_ijb1vsN}
\begin{tabular}{c|cccc|cccc}
\hline
\multirow{2}{*}{Methods  (\%) }  &  \multicolumn{4}{c}{IJB-B} &  \multicolumn{4}{c}{IJB-C} \\
\cmidrule{2-5} \cmidrule{6-9}
&FPIR=$0.01$ & FPIR=$0.1$ & Rank~$1$ & Rank~$5$ 
&FPIR=$0.01$ & FPIR=$0.1$ & Rank~$1$ & Rank~$5$ \\\hline

ResNet50    &  84.70   & 94.01   & 95.29  & 97.14
            &  92.87   & 95.28   & 96.52  &  97.69   \\ 
Prodpoly-ResNet50 &  {\bf 85.58}  \tiny{($\uparrow 0.88$)}   & {\bf 94.69}  \tiny{($\uparrow 0.68$)}  & {\bf 95.52}  \tiny{($\uparrow 0.23$)}  & {\bf 97.16}  \tiny{($\uparrow 0.02$)} 
                  &  {\bf 93.60}  \tiny{($\uparrow 0.73$)}  &  {\bf 95.93}  \tiny{($\uparrow 0.65$)}  & {\bf 96.86}   \tiny{($\uparrow 0.34$)} & {\bf 97.79}  \tiny{($\uparrow 0.10$)}   \\\hline 
\end{tabular} 
\end{center}
\end{table*}

\noindent {\bf Results on IJB-B and IJB-C.} The IJB-B dataset \cite{whitelam2017iarpa} contains $1,845$ subjects with $21.8$K still images and $55$K frames from $7,011$ videos. The IJB-C dataset \cite{whitelam2017iarpa} is a further extension of IJB-B, having $3,531$ subjects with $31.3$K still images and $117.5$K frames from $11,779$ videos. On IJB-B and IJB-C datasets, there are two evaluation protocols, 1:1 verification and 1:N identification.

\begin{figure}[h!]
\centering
\subfloat[ROC for IJB-B]{
\label{pic:ijbb_roc} 
\includegraphics[width=0.228\textwidth]{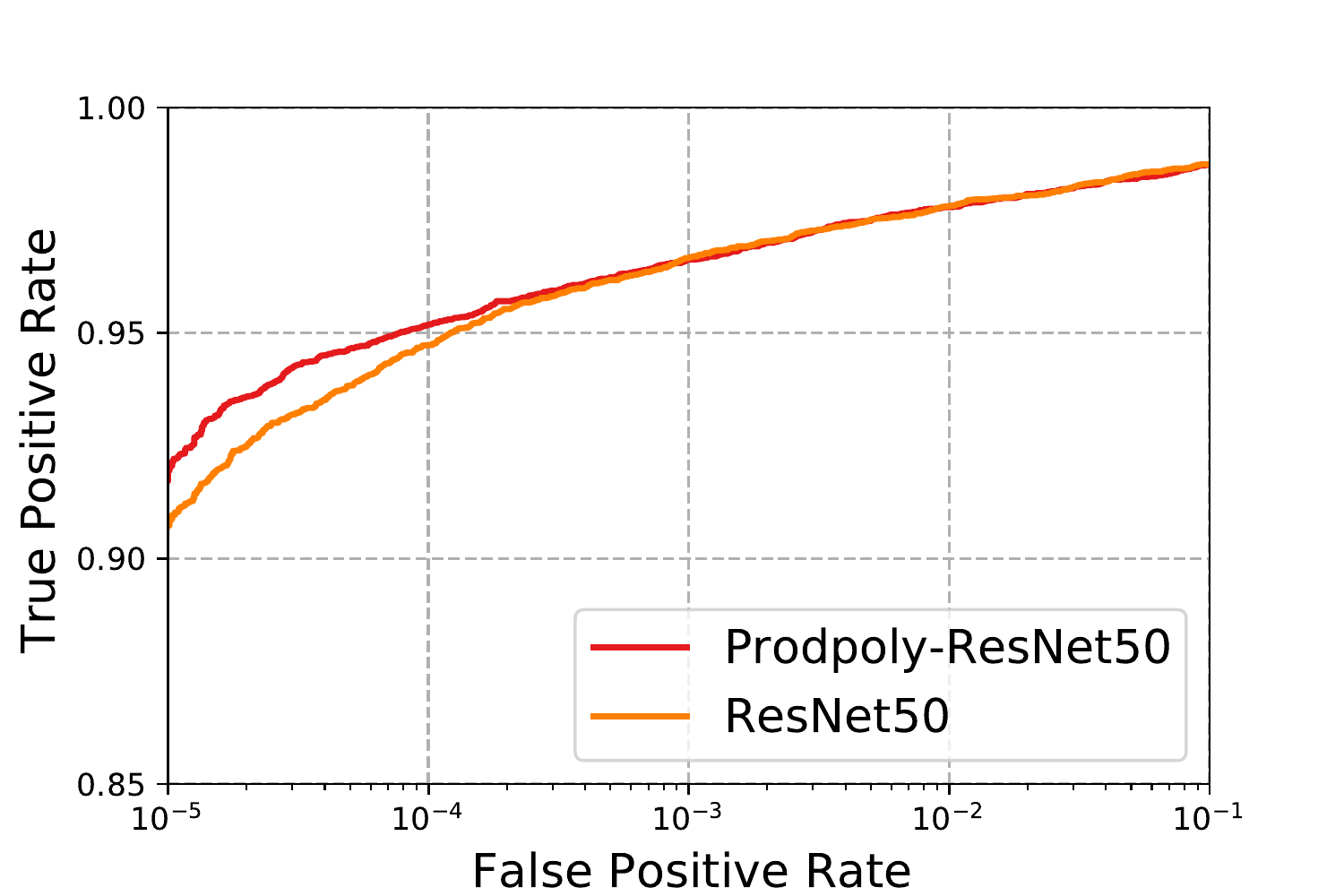}}
\subfloat[ROC for IJB-C]{
\label{pic:ijbc_roc} 
\includegraphics[width=0.228\textwidth]{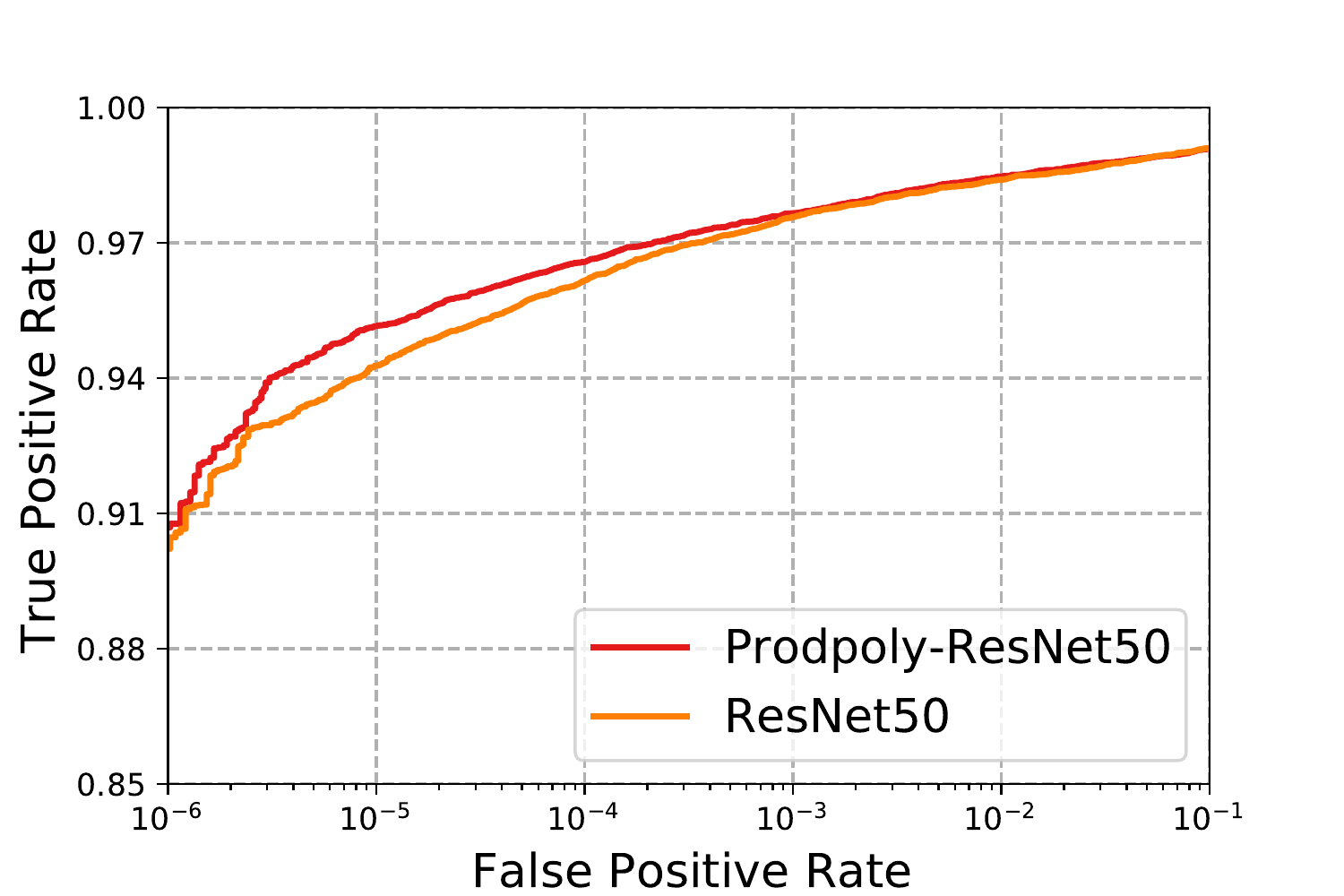}}
\caption{ROC curves of ResNet50 and Prodpoly-ResNet50 under 1:1 verification protocol on the IJB-B and IJB-C dataset.}
\label{pic:ijb}
\end{figure}

In Figure \ref{pic:ijb}, ROC curves of \resnet50 and \modelres50 under 1:1 verification protocol on IJB-B and IJB-C is plotted. 
On IJB-B, there are $12,115$ templates with $10,270$ genuine matches and $8$M impostor matches. On IJB-C, there are $23,124$ templates with $19,557$ genuine matches and $15,639$K impostor matches. The proposed method 
surpasses the baseline by a clear margin. The comparison of TAR in Table \ref{tab:comp_ijb1vs1} illustrates that \modelres50 improves the TAR (@FAR=1e-4) by $0.46\%$ and $0.41\%$ on IJB-B and IJB-C, respectively. 

Table \ref{tab:comp_ijb1vsN} compares \resnet50 and \modelres50 under the 1:N end-to-end mixed protocol, which contains both still images and full-motion videos. On IJB-B, there are $10,270$ probe templates containing $60,758$ still images and video frames.
On IJB-C, there are $19,593$ probe templates containing $127,152$ still images and video frames.
\modelres50 outperforms \resnet50 by $0.23\%$ and $0.34\%$ on IJB-B and IJB-C rank-1 face identification.

\begin{figure}[h!]
\centering
\subfloat[CMC]{
\label{fig:megafacecmc}
\includegraphics[width=0.226\textwidth]{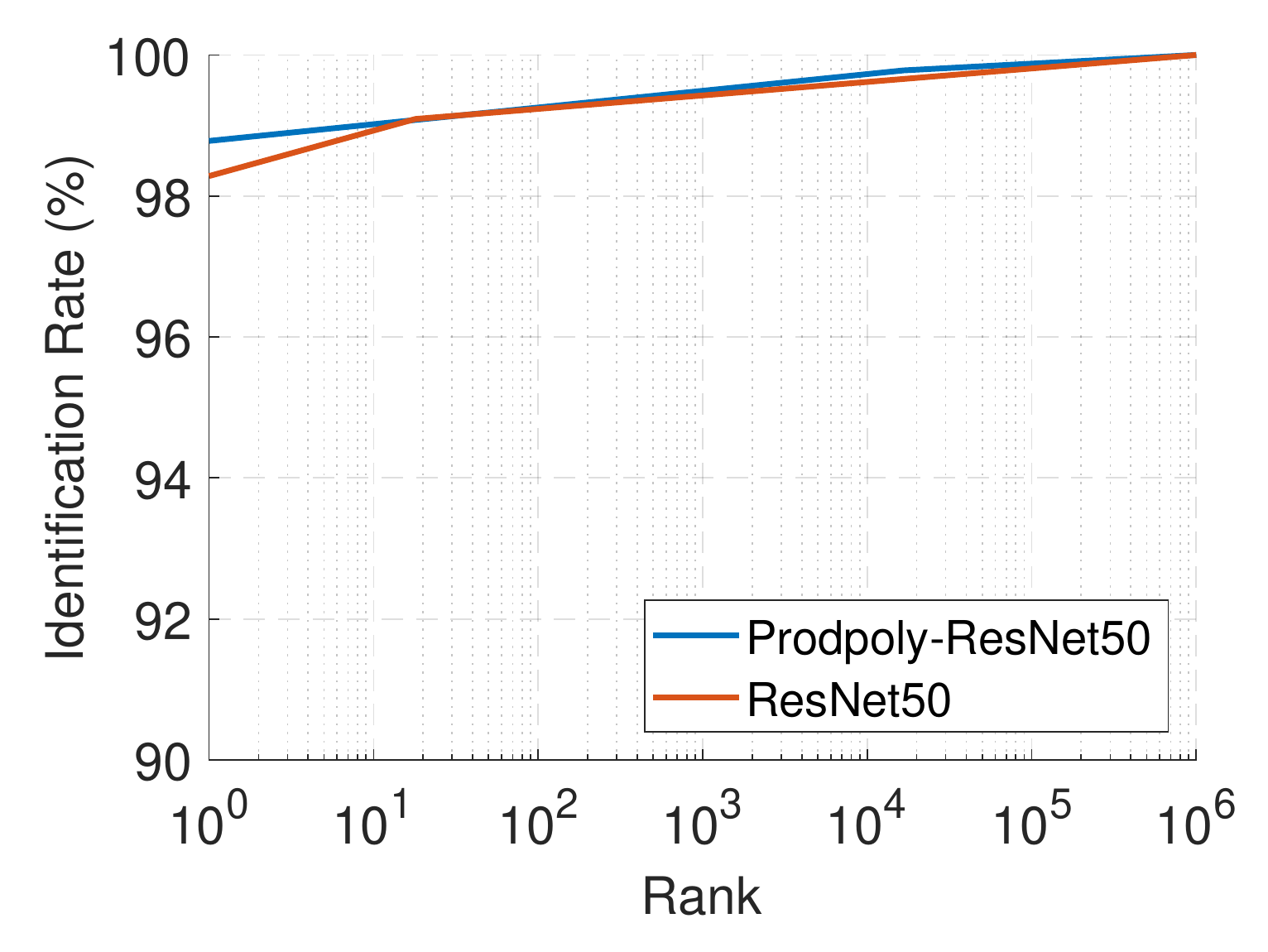}}
\subfloat[ROC]{
\label{fig:megafaceroc}
\includegraphics[width=0.226\textwidth]{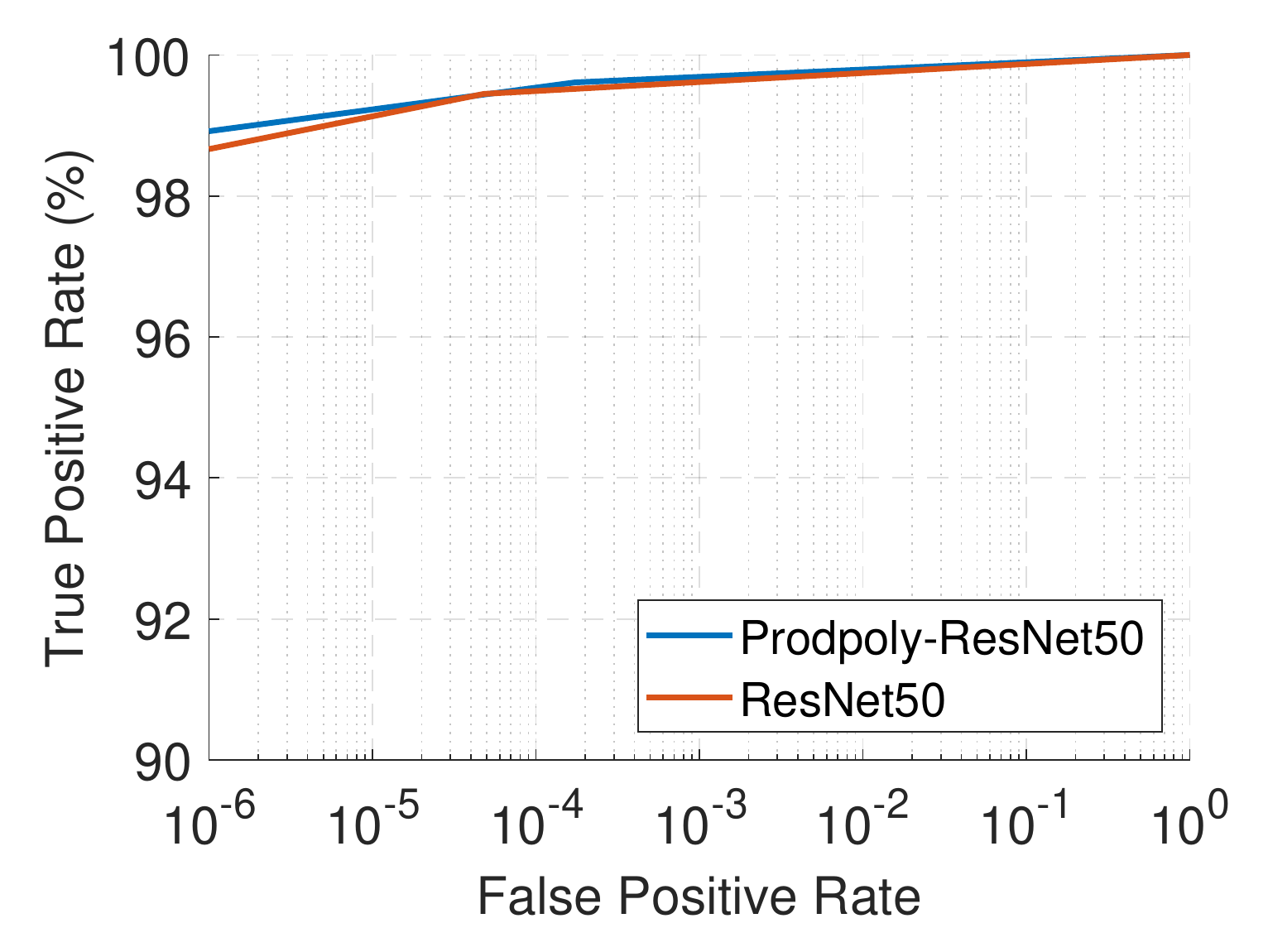}}
\caption{CMC and ROC curves of ResNet50 and the proposed Prodpoly-ResNet50 on MegaFace. Results are evaluated on the refined MegaFace dataset~\cite{deng2019arcface}.}
\label{fig:megafacecmcroc}
\end{figure}

\noindent {\bf Results on MegaFace.} The MegaFace dataset \cite{kemelmacher2016megaface} includes 1M images of 690K different individuals as the gallery set and 100K photos of $530$ unique individuals from FaceScrub \cite{ng2014data} as the probe set. On MegaFace, there are two testing protocols (e.g., identification and verification). Table \ref{table:megaface} show the identification and verification results on
MegaFace dataset. In particular, the proposed Prodpoly-ResNet50 
achieve $0.50\%$ improvement at the Rank-1@1e6 identification rate and $0.31\%$ improvement at the verification
TPR@FAR=1e-6 rate over the baseline ResNet50. In Figure \ref{fig:megafacecmcroc}, Prodpoly-ResNet50 shows superiority over ResNet50 and forms an upper envelope under both identification and verification scenarios.

\begin{table}[t!]
\begin{center}
\caption{Face identification and verification evaluation of ResNet50 and the proposed Prodpoly-ResNet50 on MegaFace Challenge1 using FaceScrub as the probe set. ``Id'' refers to the rank-1 face identification accuracy with 1M distractors, and ``Ver'' refers to the face verification TAR at $10^{-6}$ FAR. Results are evaluated on the refined MegaFace dataset~\cite{deng2019arcface}.}
\label{table:megaface}
\begin{tabular}{c|c|c}
\hline
Methods  & Id ($\%$) & Ver ($\%$) \\
\hline
ResNet50          & 98.28 & 98.64\\
Prodpoly-ResNet50 & {\bf 98.78} (\tiny{$\uparrow 0.50$)}  & {\bf 98.95} \tiny{($\uparrow 0.31$)} \\
\hline
\end{tabular}
\end{center}
\end{table}

\subsection{3D Mesh representation learning}
\label{ssec:prodpoly_mesh_representation_learning_experiment}

\begin{figure}
\centering
    \centering
    \includegraphics[width=\linewidth]{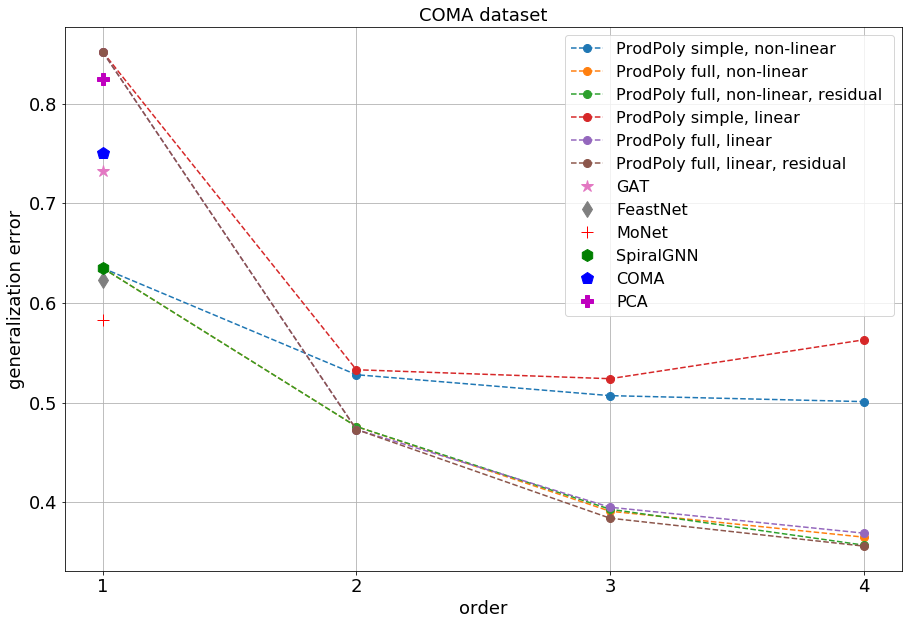}
    \includegraphics[width=\linewidth]{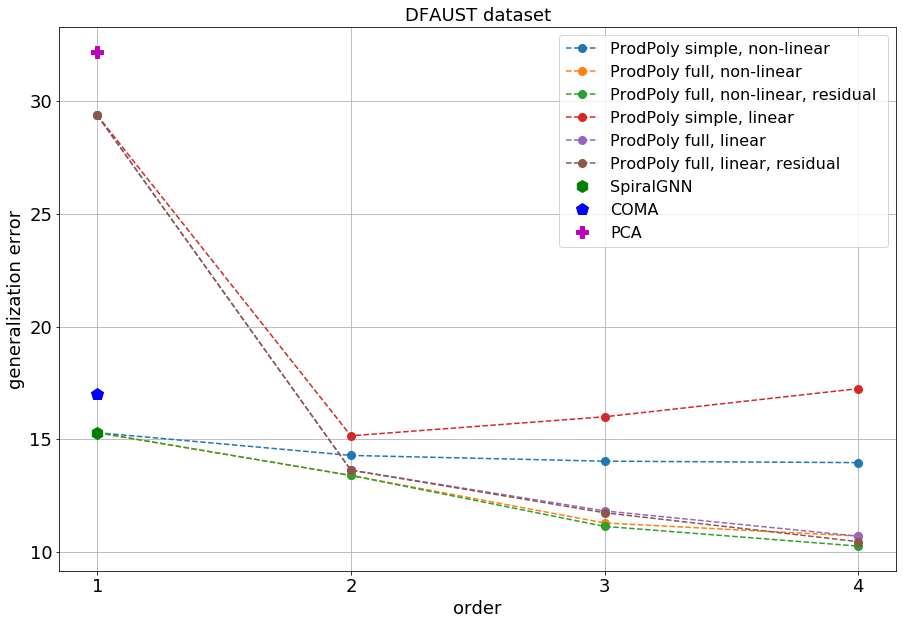}
\caption{\modelname{} vs $1^{st}$ order graph learnable operators for mesh autoencoding. Note that even without using activation functions the proposed methods significantly improve upon the state-of-the-art.}
\label{fig:prodpoly_graphs}
\end{figure}

Below, we evaluate higher order correlations in graph related tasks. We experiment with 3D deformable meshes of fixed topology~\cite{ranjan2018generating}, i.e., the connectivity of the graph $\mathcal{G}= \{\mathcal{V},\mathcal{E}\}$ remains the same and each different shape is defined as a different signal $\bm{x}$ on the vertices of the graph: $\bm{x}:\mathcal{V}\to\mathbb{R}^d$. As in the previous experiments, we extend a state-of-the-art operator, namely spiral convolutions \cite{Bouritsas_2019_ICCV}, with the \modelname{} formulation and test our method on the task of autoencoding 3D shapes. We use the existing architecture and hyper-parameters of \cite{Bouritsas_2019_ICCV}, thus showing that \modelname{} can be used as a plug-and-play operator to existing models, turning the aforementioned one into a Spiral $\Pi$-Net.
Our implementation uses a product of polynomials (referred as \textit{\modelname{} full}), where each layer is a $N^{th}$ order polynomial instantiated as a specific case of \eqref{eq:prodpoly_model2} or \eqref{eq:prodpoly_model3}: 
\newline
\textbf{\modeltwo}: ${\boutvar_{n} = \Big(\bm{A}\matnot{n}^T \boutvar_{1}\Big) * \Big(\bm{S}\matnot{n}^T \boutvar_{n-1}\Big) + \bm{A}\matnot{n}^T\boutvar_{1}}$  
\newline
\textbf{\modelthree}: ${\boutvar_{n} = \Big(\bm{A}\matnot{n}^T \boutvar_{1}\Big) * \Big(\bm{S}\matnot{n}^T \boutvar_{n-1}\Big) + \bm{A}\matnot{n}^T\boutvar_{1} + \boutvar_{n-1}}$, 
${\boutvar = \boutvar_{N} + \bm{\beta}}$
, where $\bm{A}\matnot{n}, \bm{S}\matnot{n}$ are spiral convolutions written in matrix form, $\bm{\beta}$ is a bias vector, $\boutvar_1$, $\boutvar$ is the input (which is equal to the output of the previous layer) and the output of the layer respectively. Stability of the optimization is ensured by applying vertex-wise instance normalization on the $2^{nd}$ order term of the recursive formulation.

Additionally, we evaluate our formulation with a simpler model (\textit{\modelname{} simple}) that allows for an attractive trade-off between increased expressivity and constrained parameter budget. In specific, we can create higher-order polynomials without adding new blocks in the original architecture as follows:
\newline
$\boutvar_{N} = 
\sum_{n=2}^N \underbrace{\Big(\bm{S}^T\boutvar_{1}\Big)*\Big(\bm{S}^T\boutvar_{1}\Big)\cdots\Big(\bm{S}^T\boutvar_{1}\Big)}_\text{n times} + \bm{S}^T\boutvar_{1} + \bm{\beta}$. We use the same normalization scheme as before, by independently normalizing each higher order term. Note that here we only use one learnable operator $\bm{S}$ (spiral convolution) per layer. It is interesting to notice that this model can be also re-interpreted as a learnable polynomial activation function as in \cite{kileel2019expressive}, which is a specific case of \modelname{}. 
Polynomial activation functions lead to increased expressivity per se, but are less expressive when compared to richer multiplicative interactions as introduced by our NCP and NCP-skip models. \rebuttal{In addition, as can be seen in Fig. \ref{fig:prodpoly_graphs}, experimental evidence suggests that such interactions also lead to improved empirical performance.}

\begin{figure}
\centering
    \centering
    \includegraphics[width=\linewidth]{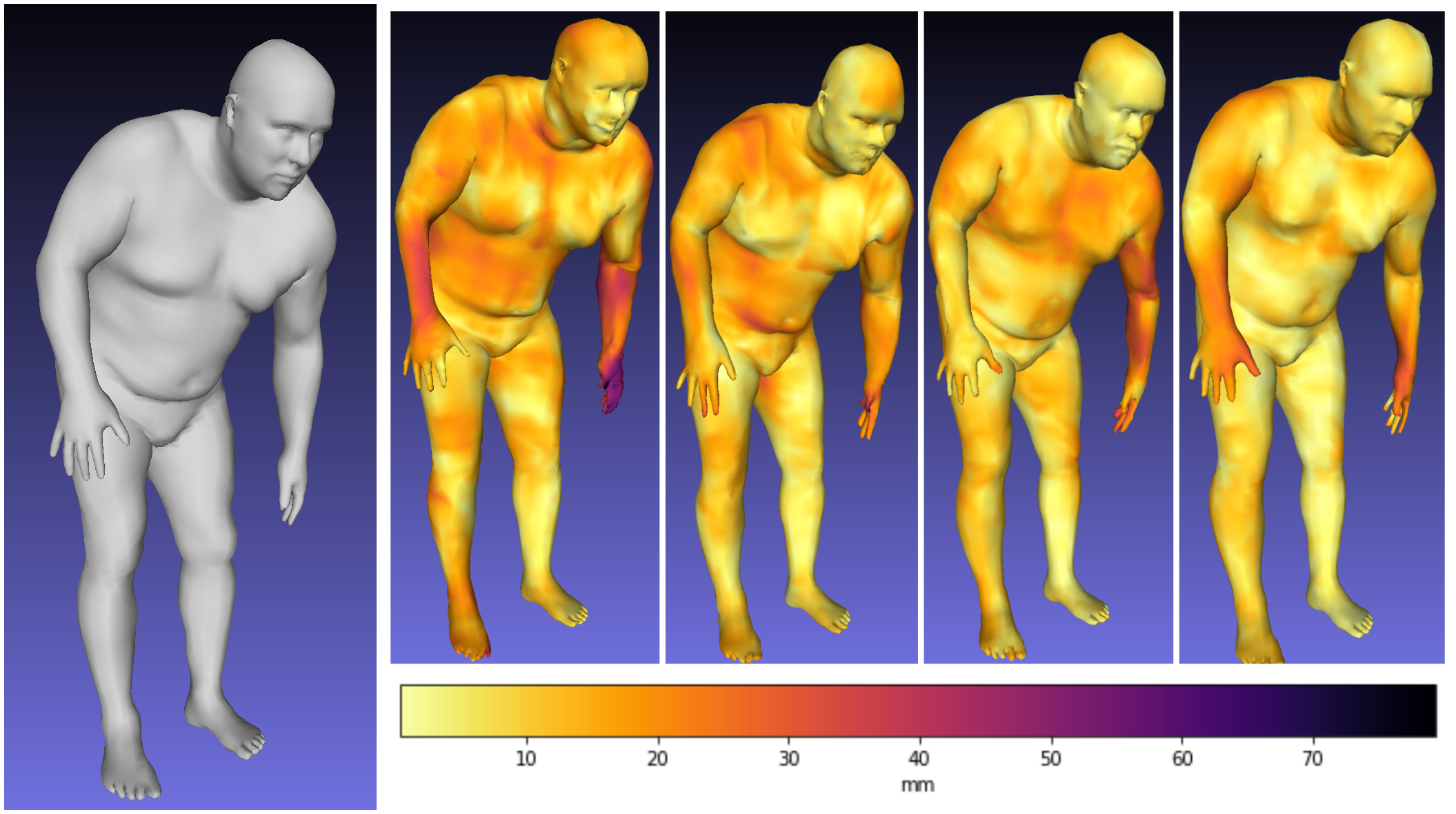}
\caption{Color coding of the per vertex reconstruction error on an exemplary human body mesh. From left to right: ground truth mesh, 1st order SpiralGNN, $2^{nd}$, $3^{rd}$ and $4^{th}$ order Spiral \modelname{}.}
\label{fig:color_coding}
\end{figure}

In Fig.~\ref{fig:prodpoly_graphs}, we compare the reconstruction error of the proposed method to the baseline spiral convolutions along with other popular graph learnable operators, i.e., the Graph Attention Network (GAT) \cite{velickovic2018graph}, FeastNet \cite{verma2018feastnet}, Mixture model CNNs (MoNet) \cite{monti2017geometric}, Convolutional Mesh Autoencoders (COMA) \cite{ranjan2018generating} which are based on the spectral graph filters of ChebNet \cite{defferrard2016convolutional}, as well as with Principal Component Analysis (PCA), which is quite popular in shape analysis applications \cite{blanz1999morphable}. The evaluation is performed on two popular 3D deformable shape benchmarks, COMA \cite{ranjan2018generating} and DFAUST \cite{bogo2017dynamic}, that depict facial expressions and body poses respectively. $\Pi$-nets outperform all published methods even when discarding the activation functions across the entire network. Similar patterns emerge in both datasets: \modeltwo{} and \modelthree{} behave similarly regardless of the absence of activation functions or not, leading to an increased performance when the order of the polynomial increases, i.e. \rebuttal{empirical performance} improves along with expressivity. Moreover, the simple model provides a boost in performance as well, although we observe a decrease for the $3^{rd}$ and $4^{th}$ order of the linear model, which might be attributed to overfitting (similarly to the linear experiments in \rebuttal{Sec. 5 in the supplementary material}). Overall, we showcase that \rebuttal{performance} may seamlessly improve by converting the existing architecture to a polynomial, without having to increase the depth or width of the architecture as frequently done by ML practitioners, and with small sacrifices in terms of inference time (\rebuttal{see Sec. 6, supplementary}) and parameter count.

Finally, in Fig.~\ref{fig:color_coding} we assess how the order of the polynomial qualitatively reflects in the reconstruction of an exemplary mesh. In particular, we color code the per vertex reconstruction error on the reconstructed meshes (right) and compare them with the input (left). Notice that the overall shape resembles the input more as we increase the order of the polynomial (especially in the head), while body parts with strong articulations (e.g. hands) are reconstructed with higher fidelity.

 \section{Future directions}
\label{sec:prodpoly_discussion}

The new class of $\Pi-$nets has strong experimental results and few empirical theoretical results already. We expect in the following years new works that improve our results and extend our formulation. To that end, we summarize below several fundamental topics that are open for interested practitioners.

The generalization of the $\Pi-$nets is a crucial topic. In our evaluation without activation functions, we noticed that polynomials might be prone to overfitting (e.g., in the classification setting without activation functions in the supplementary). When we add the non-linear activation functions we did not observe such a consistent pattern. 

In this work, we created a link between different decompositions and the resulting architectures (the three decompositions resulted in three different architectures). The relationship between neural architecture search and the tensor decomposition can be further nurtured.

Reducing the network redundancy is also an exciting topic. The theoretical properties of multiplicative interactions along with our experiments, exhibit how polynomial neural networks can be used to reduce the network redundancy. Additional post-processing techniques, such as pruning, or exploiting tools from the tensor methods, such as low-rank constraints, might be beneficial in this context. 

\rebuttal{Lastly, $\Pi-$nets inherit the properties of polynomials, e.g., higher-order terms might result in unbounded gradients. That makes studying normalization schemes of paramount significance. There might be normalization techniques that obtain a superior performance to the batch/instance normalization we employed.}
 \section{Conclusion}
\label{sec:prodpoly_conclusion}

In this work, we have introduced a new class of DCNNs, called $\Pi$-Nets, that perform function approximation using a polynomial neural network. Our $\Pi$-Nets can be efficiently implemented via a special kind of skip connections that lead to high-order polynomials, naturally expressed with tensorial factors. The proposed formulation extends the standard compositional paradigm of overlaying linear operations with activation functions. We motivate our method by a sequence of experiments without activation functions that showcase the expressive power of polynomials, and demonstrate that $\Pi$-Nets are effective in both discriminative, as well as generative tasks. Trivially modifying state-of-the-art architectures in image generation, image and audio classification, face verification/identification as well as mesh representation learning, the performance consistently improves.

 \section{Acknowledgements}
\label{sec:prodpoly_acks}
GC conducted this work while at Imperial College London. 
The work of SM, and GB was partially funded by an Imperial College DTA. The work of JD was partially funded by Imperial President's PhD Scholarship. The work of SZ was partially funded by the EPSRC Fellowship DEFORM: Large Scale Shape Analysis of Deformable Models of Humans (EP/S010203/1) and a Google Faculty Award.

{\small
\bibliographystyle{IEEEtran}
\bibliography{egbib}
}

\begin{IEEEbiography}[{\includegraphics[width=1in,height=1.25in,clip,keepaspectratio]{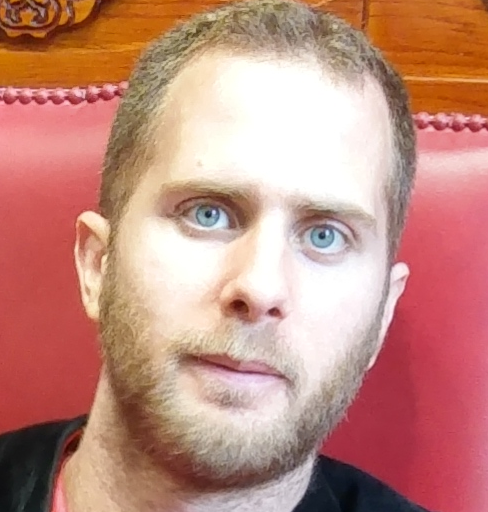}}]{Grigorios G. Chrysos}
is a Post-doctoral researcher at Ecole Polytechnique Federale de Lausanne (EPFL) following the completion of his PhD at Imperial College London (2020). Previously, he graduated from National Technical University of Athens with a Diploma/MEng in Electrical and Computer Engineering (2014). He has published his work on deformable models in prestigious journals (T-PAMI, IJCV, T-IP), while he has co-organised workshops for deformable models, e.g. 2D/3D facial landmark tracking, in CVPR/ICCV. He is a reviewer in prestigious journals including T-PAMI, IJCV, and in top tier conferences. Currently, his primary research interest is on machine learning, including generative models, tensor decompositions and modelling high dimensional distributions; his recent work has been published in top tier conferences (CVPR, ICML, ICLR).\looseness-1
\end{IEEEbiography}

\begin{IEEEbiography}[{\includegraphics[width=1in,height=1.25in,clip,keepaspectratio]{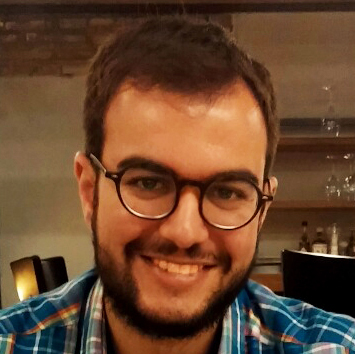}}]{Stylianos Moschoglou}
received his Diploma/MEng in Electrical and Computer Engineering from Aristotle University of Thessaloniki, Greece, in 2014. In 2015-16, he pursued an MSc in Computing (specialisation Artificial Intelligence) at Imperial College London, U.K., where he completed his project under the supervision of Dr. Stefanos Zafeiriou.
He is currently a PhD student at the Department of Computing, Imperial College London, under the supervision of Dr. Stefanos
Zafeiriou. His interests lie within the area of Machine Learning and in particular in Generative Adversarial Networks and Component Analysis.\looseness-1
\end{IEEEbiography}

\begin{IEEEbiography}[{\includegraphics[width=1in,height=1.25in,clip,keepaspectratio]{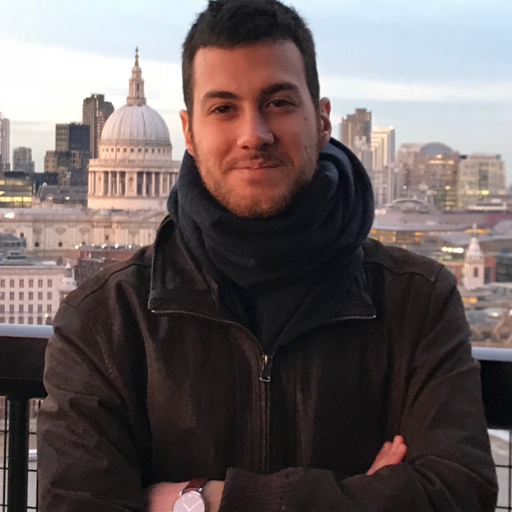}}]{Giorgos Bouritsas}
 is a PhD student at Imperial College London working with Prof. Michael Bronstein and Prof. Stefanos Zafeiriou. Giorgos graduated from National Technical University of Athens (NTUA) with an MEng Diploma in Electrical and Computer Engineering in 2017. He has spent time as a visiting researcher at the  Universitat Politècnica de Catalunya (UPC), Barcelona, as a research associate at the National Center for Scientific Research “Demokritos”, Athens, and as visiting PhD student at KU Leuven and École Polytechnique Fédérale de Lausanne (EPFL), conducting research on a variety of topics in computer vision and machine learning, 
 His research interests lie within the fields of on non-Euclidean deep learning, machine learning on graphs, deep learning theory and applications to network science and computer vision. 
Currently he is particularly focused on the theoretical underpinnings of graph neural networks and on generative models for non-Euclidean data. \looseness-1
\end{IEEEbiography}

\begin{IEEEbiography}[{\includegraphics[width=1in,height=1.25in,clip,keepaspectratio]{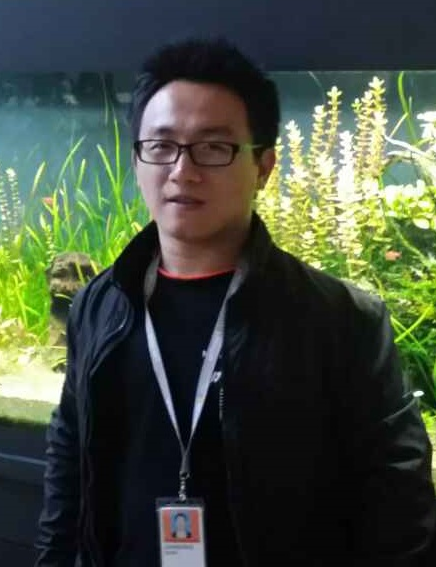}}]{Jiankang Deng} is a Ph.D. candidate in the Intelligent Behaviour Understanding Group (IBUG) at Imperial College London (ICL), supervised by Stefanos Zafeiriou and funded by the Imperial President's PhD Scholarships. He is in the project of EPSRC FACER2VM (Face Matching for Automatic Identity Retrieval, Recognition, Verification and Management). His Ph.D. research topic is face analysis (face detection, face alignment, face recognition and face generation). During his PhD studies, he has organised the Menpo 2D Challenge (CVPR 2017), the Menpo 3D Challenge (ICCV 2017) and Lightweight Face Recognition Challenge (ICCV 2019). He also won many academic challenges, such as ILSVRC Object Detection and Tracking 2017, Activity-Net Untrimmed Video Classification 2017, iQIYI Celebrity Video Identification Challenge 2018, Disguised Face Recognition Challenge 2019. He is a reviewer in prestigious computer vision journals and conferences including T-PAMI, IJCV, CVPR, ICCV and ECCV. He is the main contributor of the widely used open-source platform Insightface. 
\end{IEEEbiography}

\begin{IEEEbiography}[{\includegraphics[width=1in,height=1.25in,clip,keepaspectratio]{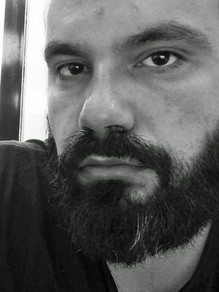}}]{Yannis Panagakis} is an Associate Professor  of machine learning and signal processing at the University of Athens. His research interests lie in machine learning and its interface with signal processing, high-dimensional statistics, and computational optimization. Specifically, Yannis is working on models and algorithms for robust and efficient learning from high-dimensional data and signals representing audio, visual, affective, and social information. He has been awarded the prestigious Marie-Curie Fellowship, among various scholarships and awards for his studies and research. He co-organized the BMVC 2017 conference and several workshops and special sessions in top venues such as ICCV. He received his PhD and MSc degrees from the Department of Informatics, Aristotle University of Thessaloniki and his BSc degree in Informatics and Telecommunication from the University of Athens, Greece.
\end{IEEEbiography}

\begin{IEEEbiography}[{\includegraphics[width=1in,height=1.25in,clip,keepaspectratio]{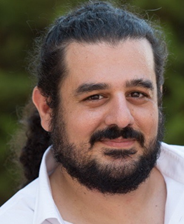}}]{Stefanos Zafeiriou}
(M’09) is a Professor in Machine Learning and Computer Vision with the Department of Computing, Imperial College London, U.K, and a Distinguishing Research Fellow with University of Oulu. He was a recipient of the Prestigious Junior Research Fellowships from Imperial College London in 2011 to start his own independent research group. He was the recipient of the President’s Medal for Excellence in Research Supervision for 2016. He currently serves as an Associate Editor of the IEEE Transactions on Affective Computing and Computer Vision and Image Understanding journal. He has been a Guest Editor of over six journal special issues and co-organised over 13 workshops/special sessions on specialised computer vision topics in top venues, such as CVPR/FG/ICCV/ECCV. He has co-authored over 55 journal papers mainly on novel statistical machine learning methodologies applied to computer vision problems, such as 2-D/3-D face analysis, deformable object fitting and tracking, published in the most prestigious journals in his field of research, such as the IEEE T-PAMI, the International Journal of Computer Vision, the IEEE T-IP, the IEEE T-NNLS, the IEEE T-VCG, and the IEEE T-IFS, and many papers in top conferences. He has more than 14,000 citations to his work, h-index 57.\looseness-1
\end{IEEEbiography} 
\end{document}